\documentclass[twoside]{article}

\usepackage[accepted]{aistats2023}
\usepackage[round]{natbib}
\usepackage{amsmath,amsfonts,bm}

\newcommand{\x}{\mathbf{x}}
\newcommand{\y}{\mathbf{y}}

\newcommand{\X}{\mathbf{X}}

\newcommand{\W}{\mathbf{W}}
\newcommand{\w}{\mathbf{w}}
\newcommand{\U}{\mathbf{u}}
\newcommand{\h}{\mathbf{H}}
\newcommand{\xij}{\hat{\mathbf{x}}_i^{(j)}}
\newcommand{\paren}[1]{\left(#1\right)}
\newcommand{\norm}[1]{\left\|#1\right\|}
\newcommand{\indy}[1]{\mathbb{I}\left\{#1\right\}}

\newcommand{\inner}[2]{\left\langle#1, #2\right\rangle}

\def\figref#1{figure~\ref{#1}}
\def\eqref#1{equation~\ref{#1}}
\def\1{\bm{1}}

\DeclareMathAlphabet{\mathsfit}{\encodingdefault}{\sfdefault}{m}{sl}
\SetMathAlphabet{\mathsfit}{bold}{\encodingdefault}{\sfdefault}{bx}{n}

\newcommand{\E}{\mathbb{E}}

\newcommand{\R}{\mathbb{R}}

\usepackage{microtype}
\usepackage{graphicx}
\usepackage{booktabs}
\usepackage{hyperref}

\usepackage{amsmath}
\usepackage{amsthm}
\usepackage{algorithm}
\usepackage{algorithmic}
\usepackage{wrapfig}
\usepackage{multirow}
\usepackage{amssymb}
\usepackage{xfrac}
\usepackage{enumitem}
\usepackage{caption}
\usepackage{subcaption}
\usepackage{zref-xr}
\usepackage{zref-user}

\usepackage[absolute,overlay]{textpos}
\usepackage{tikz}

\theoremstyle{definition}

\newtheorem{theorem}{Theorem}
\newtheorem{corollary}{Corollary}
\newtheorem{lemma}{Lemma}

\newtheorem{asump}{Assumption}

\begin{document}

\twocolumn[

\aistatstitle{\textsc{LoFT}: Finding Lottery Tickets through Filter-wise Training}

\aistatsauthor{Qihan Wang$^\star$ \And Chen Dun$^\star$ \And  Fangshuo Liao$^\star$ \AND Chris Jermaine \and Anastasios Kyrillidis }

\aistatsaddress{Rice University} ]

\begin{abstract}
\vspace{-0.2cm}
Recent work on the Lottery Ticket Hypothesis (LTH) shows that there exist ``\textit{winning tickets}'' in large neural networks. 
These tickets represent ``sparse'' versions of the full model that can be trained independently to achieve comparable accuracy with respect to the full model. 
However, finding the winning tickets requires one to  \emph{pretrain} the large model for at least a number of epochs, which can be a burdensome task, especially when the original neural network gets larger. 

In this paper, we explore how one can efficiently identify the emergence of such winning tickets, and use this observation to design efficient pretraining algorithms.
For clarity of exposition, our focus is on convolutional neural networks (CNNs). %, which are more complex than simple multi-layer perceptrons, but simple enough to exposure our ideas. 
To identify good filters, we propose a novel filter distance metric that well-represents the model convergence.
% I think here we need to know the true winning tickets in order to evaluate a ticket using this filter distance.
As our theory dictates, our filter analysis behaves consistently with recent findings of neural network learning dynamics.
Motivated by these observations, we present the \emph{LOttery ticket through Filter-wise Training} algorithm, dubbed as \textsc{LoFT}.
\textsc{LoFT} is a model-parallel pretraining algorithm that partitions convolutional layers by filters to train them independently in a distributed setting, resulting in reduced memory and communication costs during pretraining.
Experiments show that \textsc{LoFT} $i)$ preserves and finds good lottery tickets, while $ii)$ it achieves non-trivial computation and communication savings, and maintains comparable or even better accuracy than other pretraining methods.
% This part is sort of repeating "preserves and finds good lottery tickets".
\end{abstract}

\vspace{-0.2cm}
\section{Introduction}
\label{sec:intro}
\vspace{-0.2cm}
The Lottery Ticket Hypothesis (LTH) \cite{frankle2018lottery} claims that neural networks (NNs) contain subnetworks (``\emph{winning tickets}'') that can match the dense network's performance when fine-tuned in isolation. 
Yet, identifying such subnetworks often requires proper pretraining of the dense network.
Empirical studies based on this statement show that NNs could potentially be significantly smaller without sacrificing accuracy
%---this observation is based mostly on empirical studies
\cite{pretrn_lth,stable_lth,state_of_sparsity,rethinking-pruning,one_ticket_wins,deconstructing_lth,to_prune_or_not}.\footnote{With the exception of \cite{pruning_is_all_you_need,log_pruning_is_all_you_need,subset_lth} that focus on finding subnetworks from randomly initialized NNs without formal training.}
How to efficiently find such subnetworks remains a widely open question: since LTH relies on a \emph{pretraining} phase, it is a \emph{de facto} criticism that finding such pretrained models could be a burdensome task, especially when one focuses on large NNs. 

This burden has been eased with efficient training methodologies, which are often intertwined with pruning steps. 
Simply put, one has to answer two fundamental questions: \emph{``When to prune?''} and \emph{``How to pretrain such large models?''}. \
Focusing on \emph{``When to prune?''}, one can prune before \cite{lee2018snip,lee2019signal,wang2019picking}, after \cite{lecun1990optimal,hassibi1993optimal,dong2017learning,han2015learning,li2016pruning,molchanov2019pruning,han2015deep,wang2019eigendamage,zeng2019mlprune}, and/or during pretraining \cite{frankle2018lottery,srinivas2016generalized,louizos2018learning,bellec2018deep,dettmers2019sparse,mostafa2019parameter,mocanu2018scalable}.\footnote{LTH approaches, while they originally imply pruning after training, include pruning at various stages during pretraining to find the sparse subnetworks.}
Works like SNIP \cite{lee2018snip,lee2019signal} and GraSP \cite{wang2019picking} aim to prune without pretraining, while suffering some accuracy loss.
% For all cases, the existing literature has shown relatively good performance---as in preserving the winning tickets---that complements the pruning family. 
% Roughly, pruning at initialization saves resources both during testing and training time, though with a drop in accuracy \cite{wang2019picking}.
Pruning after training often leads to favorable accuracy, with the expense of fully training a large model.
A compromise between the two approaches exist in \emph{early bird tickets} \cite{early-bird}, where one could potentially avoid the full pretraining cost, but still identify ``winning tickets'', by performing a smaller number of training epochs and lowering the precision of computations.
\textit{This suggests the design of more efficient pretraining algorithms that target specifically at identifying the winning tickets for larger models.}
% (delete?)Focusing on \emph{``How often to prune?''}, the answer to this question has led to a series of empirical hyperparameter schedules.
% There exist iterative procedures, such as the Iterative Magnitude Pruning (IMP) \cite{frankle2018lottery}, that prune only a small number of weights per epoch, and re-enter the \texttt{train/score/prune/rewind} cycle. 
% While such techniques show excellent performance \cite{frankle2018lottery}, this is a result of heavy hyperparameter tuning: e.g., setting the pruning rate per iteration, using adaptive learning rates and potentially a warmup phase, as well as setting how many total pruning iterations we need \cite{stabalizing-LTH,renda2019comparing}, among others.

Focusing on \emph{``How to pretrain large models?''}, modern large-scale neural networks come with significant computational and memory costs. 
% \emph{``How can one pretrain large models to find good tickets?''}
Researchers often turn to distributed training methods, such as data parallel and model parallel \cite{parallelsgd,parallelsgdanalysis,localsgdconverge,parallelism_survey,lamp,integrated_model_parallel,xpipe,multi_gpu_model_parallel}, to enable heavy pretraining towards finding winning tickets, by using clusters of compute nodes.
Yet, data parallelism needs to update the whole model on each worker---which still results in a large memory and computational cost. 
To handle such cases, researchers utilize model parallelism, such as Gpipe \cite{huang2019gpipe}, to reduce the per node computational burden. 
Traditional model parallelism enjoys similar convergence behaviour as centralized training, but needs to synchronize at every training iteration to exchange intermediate activations and gradient information between workers, thus often incurring high communication cost. 
%\bluetxt{However, recent work on early-bird tickets \cite{early_bird} implies that convergence to a small training error may not be necessary for the discovery of winning tickets.}

% Thus, the question how to efficiently pretrain large models that preserve lottery tickets remains widely open. %distributed protocols 

\noindent \textbf{Our approach and contributions.}
We propose a new model-parallel pretraining method on the one-shot pruning setting that can efficiently reveal winning tickets for CNNs. 
In particular, we center on the following questions: %\vspace{-0.2cm}
\begin{center}
\noindent \textit{``What is a characteristic of a good pretrained CNN that contains the winning ticket? How will such a criterion inform our design towards efficient pretraining?''}
\end{center}

% \textit{Our focus is on CNNs.}
Prior works show that filter-wise pruning is more preferable compared to weight pruning for CNNs \cite{huang2019gpipe,He2020LearningFilterPruning, dacunha2022ProvingSLTH, Wang2021CNNPruning, efficient-convnet-pruning}.
Our approach operates by decomposing the full network into narrow subnetworks via filter-wise partition during pretraining.
These subnetworks --which are randomly recreated intermittently during the pretraining process-- are trained independently, and their updates are periodically aggregated into the global model. % to pretrain the full network. 
Because each subnetwork is much smaller than the full model, our approach enables scaling beyond the memory limit of a single GPU. 
Our methodology allows the discovery of winning tickets with less memory and a lower communication budget.
The contributions are summarized as follows: \vspace{-0.3cm}
\begin{itemize}[leftmargin=*]
    \item We propose a metric to quantify the distance between tickets in different stages of pretraining, allowing us to characterize the convergence to winning tickets throughout the pretraining process. \vspace{-0.1cm}
    \item We identify that such convergence behavior suggests an alternative way of pretraining: we propose a novel model-parallel pretraining method through filter-wise partition of CNNs and iterative training of such subnetworks. \vspace{-0.1cm}
    \item We theoretically show that our proposed method achieves CNN weight that is close to the weight found by gradient descent in a simplified scenario. \vspace{-0.1cm}
    \item We empirically show that our method provides a better or comparable winning ticket, while being memory and communication efficient.
\end{itemize}

\vspace{-0.3cm}
\section{Preliminaries}
\vspace{-0.2cm}
The CNN model \cite{he2015deep,NIPS2012_c399862d,simonyan2015deep} is composed of convolutional layers, batch norm layers \cite{ioffe2015batch}, pooling layers, and a final linear classifier layer. 
Our goal is to retrieve a \emph{structured} winning ticket, through partitioning and pruning the filters in the convolutional layers.

Mathematically, we formulate this process as follows.
Let $p_i$ denote the number of input channels for the $i$-th convolutional layer. 
Correspondingly, the output channel of the $i$-th layer is the same as the input channel of the $(i+1)$-th layer, which is $p_{i+1}$. 
Let $h_i$,$w_i$ be the height and width of the input feature maps, respectively. 
Then, the $i$-th convolutional layer transforms the input feature map $x_i \in \mathbb{R}^{p_i \times h_i \times w_i}$ into the output feature map $x_{i+1} \in \mathbb{R}^{p_{i+1} \times h_{i+1} \times w_{i+1}}$ by performing 2D convolutions on the input feature map with $p_{i+1}$ filters of size $3\times3$, where the $j$-th filter is denoted as $\mathcal{F}_{i,j} \in \mathbb{R}^{p_{i}\times 3 \times 3}$. 
Thus the total filter weight for the $i$-th layer is $\mathcal{F}_i \in \mathbb{R}^{p_{i+1} \times p_i \times 3 \times 3}$. 
Formally, prunning $\sfrac{1}{k}$ of the filters in the $i$-th layer is equivalent to discarding $\sfrac{p_{i+1}}{k}$ filters. 
Thus the resulted total pruned filter weight is in $\mathbb{R}^{\sfrac{p_{i+1} \cdot (k-1)}{k} \times p_i \times 3 \times 3}$ and the output feature map $x_{i+1}$ is in $\mathbb{R}^{\sfrac{p_{i+1} \cdot (k-1)}{k} \times h_{i+1} \times w_{i+1}}$. 

\vspace{-0.2cm}
\section{Identifying Tickets Early in Training}
\vspace{-0.2cm}
% \cite{early_bird} observed that winning tickets can emerge at early stages in training. 
% For CNNs, \cite{li2016pruning} proposes the method of filter-wise pruning. This leads to key questions in early ticket identification: %\vspace{-0.3cm}
In this section, we aim at answering the following questions to motivate the design of an efficient pretraining algorithm: \vspace{-0.1cm}
\begin{center}
    \textit{``How do we compare different winning filters? How early can we observe winning filters?"}
    \vspace{-0.1cm}
\end{center}
% \begin{itemize}%[leftmargin=*}
    % \item How do we compare different winning filters? %\vspace{-0.2cm} %(Section 2.1) 
    % \item How early can we observe winning filters? %\vspace{-0.2cm} %(Section 2.2)
    % \item Can we identify those winning tickets with no full training needed? (Section 2.3)
    % \item Can we find high-quality winning filters via ``cheaper'' training? %(Section 3)
% \end{itemize}
\vspace{-0.1cm}
\subsection{Evaluate the distance of two pretrained models}
\vspace{-0.1cm}
With the goal of identifying tickets early in training, we study when we can prune to find a winning ticket reliably, which oftentimes occurs before training accuracy stabilizes \cite{early_bird}. 
%In the context of pruning a CNN, that means we can prune to get a good set of filters. 
Thus, we need a metric to evaluate how trained filters evolve towards being stabilized, as the iterations increase and before they get pruned. %much the tickets trained thus far differ from each other. 
%it's natural for us look at how the winning filters changes, and use it as a distance metric. 
Since at pruning time we care about the relative magnitude of the filter weights, this question can be abstracted as \emph{finding the distance of two different rankings of a given set of filters}.

Borrowing techniques from search system rankings \cite{10.1145/1772690.1772749}, we propose a \textit{filter distance} metric based on a position-weighted version of Spearman's footrule \cite{10.2307/1422689}.
In particular, consider evaluating the distance between trained convolutional layers at epochs $X$ and $Y$. 
% \textcolor{magenta}{Maybe we can describe what we mean by ``two models''. I assume as two models we mean different trained models within the same training procedure: e.g., the same CNN on iteration i and j are two different models, right?} 
Denote their the filters at epochs $t_1$ and $t_2$ on the $i$-th layer as $\mathcal{F}^{(t_1)}_i, \mathcal{F}^{(t_2)}_i$. 
% For each of these $p_{i+1}$ filters, w
We calculate the $\ell_2$-norm of $\mathcal{F}^{(t_1)}_{i, j}, \mathcal{F}^{(t_2)}_{i, j}$ for each filter index $j\in[p_{i+1}]$ and sort them by magnitude.
% \textcolor{magenta}{Write down the $\ell_2$-norm expression. Also, it is not clear what $R^{(X)}$ and $R{(Y)}$ are.}
We denote the two sorted list with length $p_{i+1}$ as $R^{(t_1)}$ and $R^{(t_2)}$.
Each of these lists contains the $\ell_2$-norm of the filters, namely $\left\|\mathcal{F}^{(t_1)}_{i, j}\right\|_2$ and $\left\|\mathcal{F}^{(t_2)}_{i, j}\right\|_2$.
% This leads to $R^{(X)}, R^{(Y)}$: two sorted lists of $p_{i+1}$ elements, containing the quantities $\left\|\mathcal{F}^{(X)}_{i, j}\right\|_2$, $\left\|\mathcal{F}^{(Y)}_{i, j}\right\|_2$ sorted for $j \in [p_{i+1}]$.

We represent the change in ranking from $R^{(t_1)}$ to $R^{(t_2)}$ as $\sigma$. 
I.e., if $x \in R^{(t_1)}$ is the $i$-th element in $R^{(t_1)}$, then, the ranking of $x$ in $R^{(t_2)}$ is denoted as $\sigma(i)$. 
The original Spearman's footrule defines the displacement of element $i$ as $|i - \sigma(i)|$, leading to the total displacement of all elements:
\begin{align*}
F(\sigma) = \sum_i |i - \sigma(i)|.\\[-20pt]
\end{align*}
Given weights $w_i$'s for the elements, the weighted displacement for element $i$ becomes $w_i \cdot \left|\sum_{j<i} w_j - \sum_{\sigma(j) < \sigma(i)} w_j\right|$, leading to the total weighted displacement as follows:
\footnote{Using other norm calculation like $\ell_2$-norm will not affect the overall characteristic of filter distance in analysis.}
\begin{align*}
F_w(\sigma) = \sum_i w_i \cdot \left(\left|\sum_{j<i} w_j - \sum_{\sigma(j) < \sigma(i)} w_j\right|\right).\\[-25pt]
\end{align*}

% \textcolor{magenta}{Tasos: I have the feeling that here you can make it more specific by defining two sets of filters (with dimensions etc etc), and define the discussion below as it is implemented in your code.}
To put emphasis on the correct ranking of the top elements, we set the position weight for the $i$-th ranking element as $1/i$. 
To further simplify calculations, we approximate $\sum_{i = 1}^n \frac{1}{i} \approx \ln(n) - \ln(1)$ where $\ln(\cdot)$ is the natural logarithm. 
The above lead to the following definition for our \textit{filter distance}:
\begin{align*}
F_{\text{filter}}(\sigma) = \sum_i \tfrac{1}{i} \cdot \left|\ln(i) - \ln(\sigma(i))\right|. \\[-25pt]
\end{align*}

For the case where the two lists of pruned filters do not contain the same elements, we can naturally define the distance when the $i$-th element is not in the other list to be $|\ln(l+1) - \ln(i)|$; $l$ is the length of the pruned filter list. This filter distance metric is fundamentally different from the mask distance proposed in \cite{early_bird}.
A detailed comparison can be found in Related Work.
We compare against these early-pruning methods in the experiments. % when we compare with early tickets drawn in training.
\begin{figure}
        \centering
        \begin{subfigure}[b]{0.31\linewidth}
             \centering
             \includegraphics[width=\linewidth]{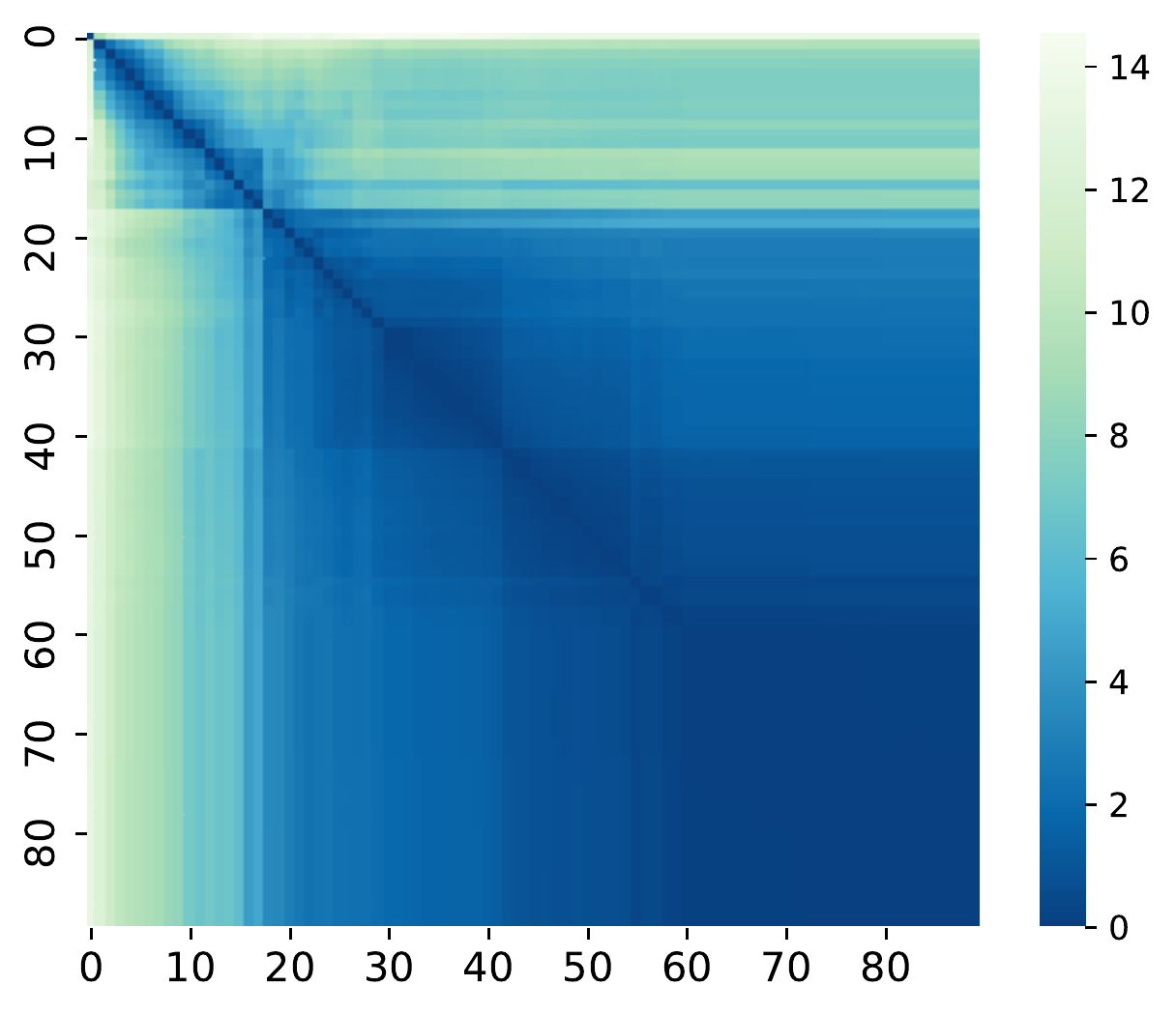}
             \caption{conv-layer2}
         \end{subfigure}
        %  \hfill
         \begin{subfigure}[b]{0.31\linewidth}
             \centering
             \includegraphics[width=\linewidth]{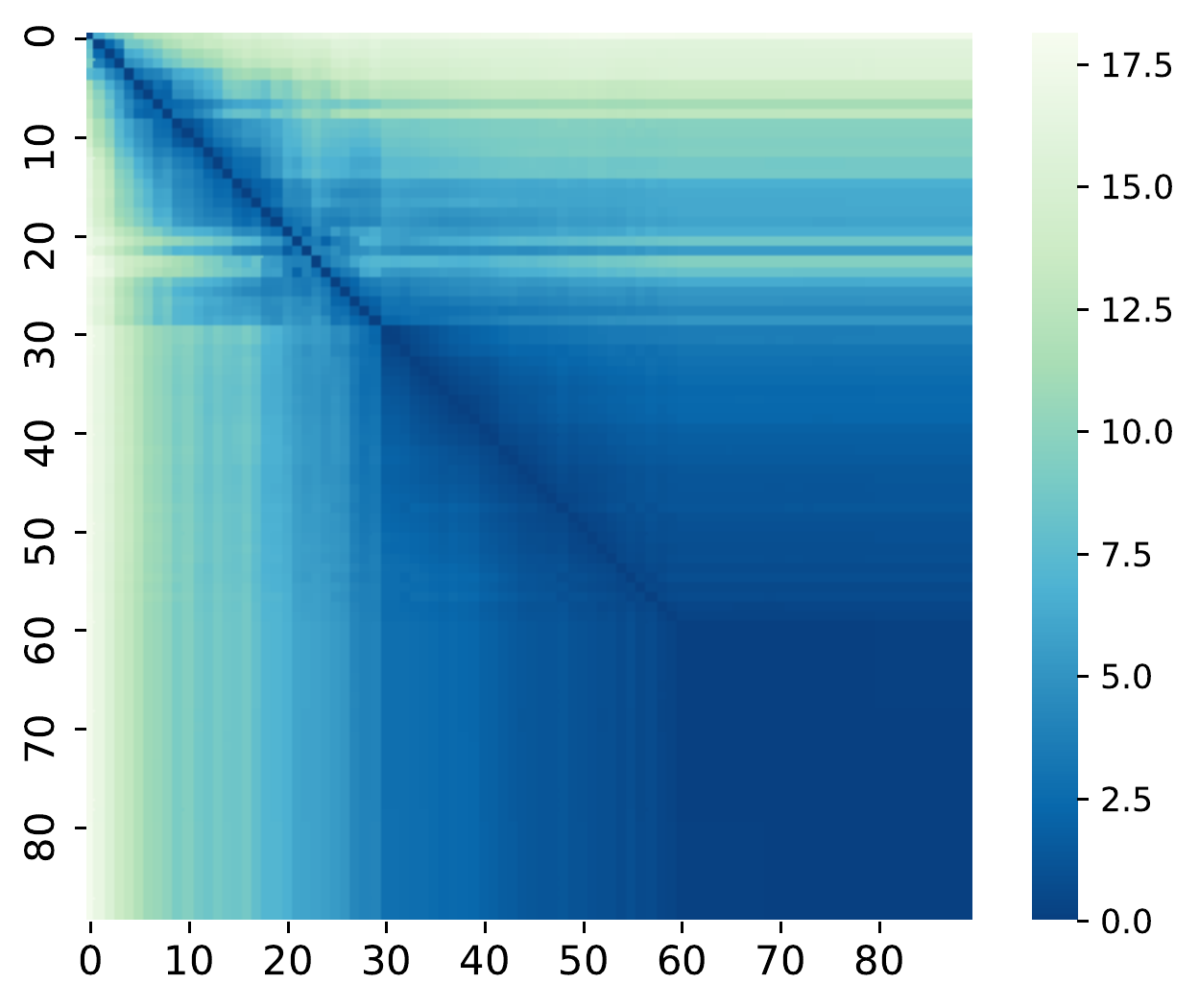}
             \caption{conv-layer3}
         \end{subfigure}
        %  \hfill
         \begin{subfigure}[b]{0.31\linewidth}
             \centering
             \includegraphics[width=\linewidth]{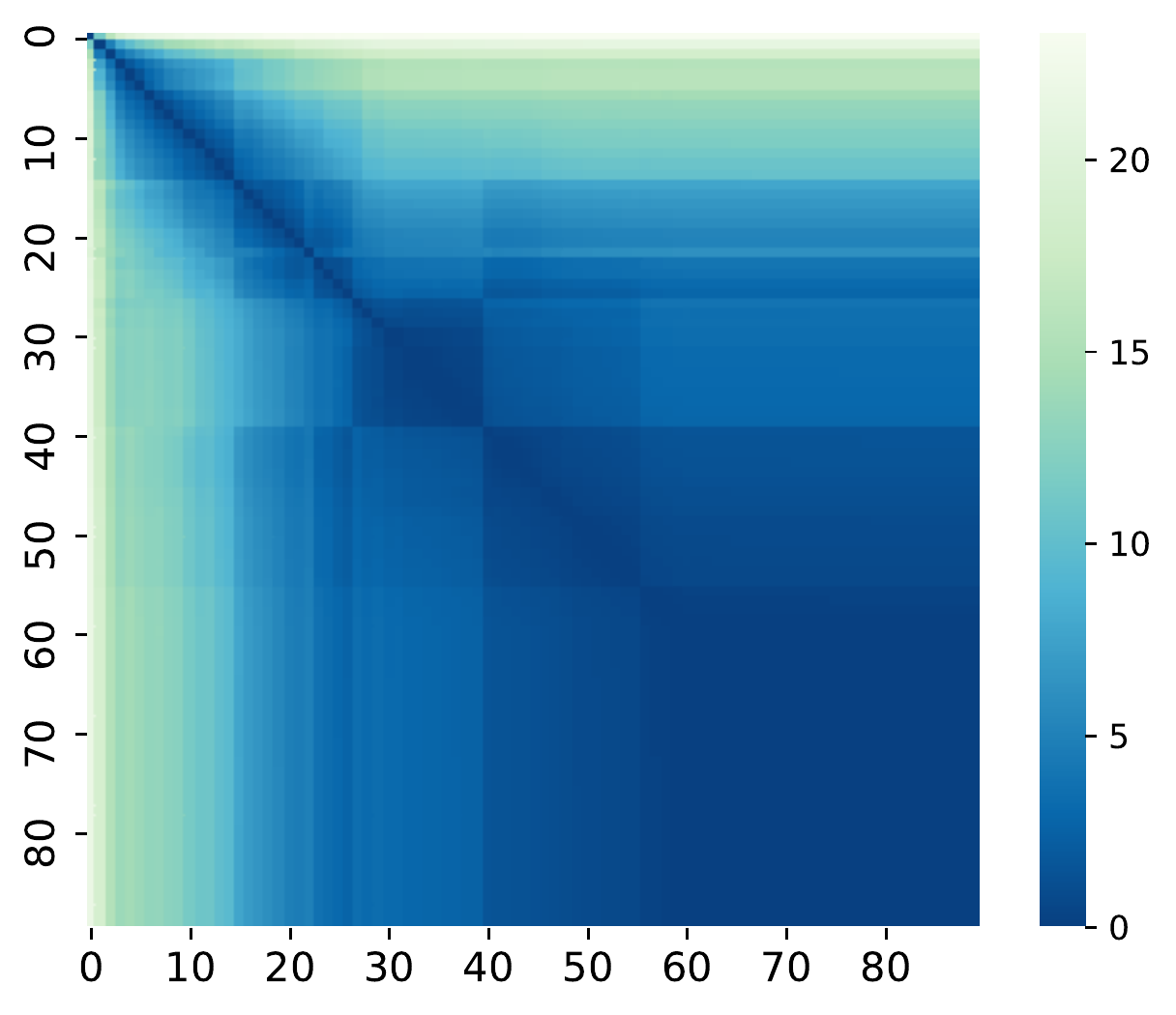}
             \caption{conv-layer4}
         \end{subfigure}
        \caption{Heatmap visualization of the pairwise mask distance for different layers of WideResNet18 trained on ImageNet. The $(i, j)$-th matrix element denotes the filter distance of a given filter between the $i$-th and $j$-th iteration. 
     Lighter color indicates a larger filter distance, while a smaller filter distance is depicted with a darker blue color.}
        \label{fig:full_heatmap}
        \vspace{-0.4cm}
\end{figure}

\vspace{-0.2cm}\subsection{How early can we observe winning filters?}
\vspace{-0.2cm}
With the \textit{filter distance} defined, we first visualize its behavior over a CNN as a function of training epochs. 
Figure \ref{fig:full_heatmap} plots the pairwise filter distance of a WideResNet18 network \cite{DBLP:journals/corr/ZagoruykoK16} on the ImageNet dataset \cite{5206848}.
Here, the $(i, j)$-th element in the heatmaps denotes the filter distance of a given filter in the network, between the $i$-th and $j$-th iteration in the experiment. 
Lighter coloring indicates a larger filter distance, while a smaller filter distance is depicted with a darker color.

Across all layers, during the first $\sim$~15 epochs, the filter distance is changing rapidly between training epochs, as is indicated by the rapidly shifting color beyond the diagonal. 
Between $\sim$~15 to $\sim$~60 epochs, filter ranking has relatively converged as the model is learning to fine-tune its weights. 
Finally, at around the $\sim$~60-th epoch, the filter distance becomes fully stable and we can observe a solid blue block at the lower right corner. 
This observation provides intuition that concur the hypothesis in \cite{achille2019critical} about a critical learning period, and observation by \cite{early_bird} using mask-distance on Batch Normalization (BN) layers.

% \subsection{Intuition behind \textsc{LoFT}}
\vspace{-0.3cm}
\subsection{Rethinking the Property of Winning Tickets}
\vspace{-0.2cm}
The empirical analysis above suggests that \textit{training the CNN weights until loss converges is not necessary for the discovery of winning tickets.}
%we could potentially find winning tickets by filter-pruning early in the training process. 
However, many existing pretraining algorithms do not exclude heavy training over the whole CNN model. 
Even though one could utilize distributed solutions with multiple workers (like the data parallel and model parallel protocols), these come with uncut computation, memory and communication costs, since these algorithms are originally designed for training to convergence. 
\textit{These facts demand a new pretraining algorithm, targeting specifically at efficiently finding winning tickets.}

Knowing the winning filters beforehand would greatly reduce the pretraining cost,
%; yet, this question is as hard as the original problem.
but this is hard to achieve in practice.
%Thus, the following question emerges:
As a compromise, we can turn to the following question:
``\textit{
Can we \textbf{randomly sample} ``tickets'' during pretraining, and  independently train them in parallel on different workers, with the hope to preserve the winning tickets?''} 
This would enable a highly efficient distributed implementation: subsets of tickets can be trained independently on each worker, with limited communication cost and less computational cost per worker.

%But in order to uncover the potential winning ticket, we need to sample and train sufficiently large number of tickets to train all the potential filters. This process can be very expensive and it is unclear how many sample tickets we need to train. 

To reduce the total computation and memory cost, one could only consider a small number of disjoint tickets per distributed worker. 
Yet, this simple heuristic should be used cautiously: in particular, \textit{splitting only once the convolutional filters --with no further communication between workers-- could miss the global winning ticket}, since no interaction is assumed between ``locally'' trained filters, leading to a strong greedy solution.
This suggests that, in order to recover a good ticket, one needs to sample and train sufficiently large number of tickets to (heuristically) assure that ``a good portion'' of filters is trained, as well as different combinations of filters are tested in each iteration.

\begin{figure*}[ht!]
    \centering
    \vspace{-0cm}
    \includegraphics[width=0.8\linewidth]{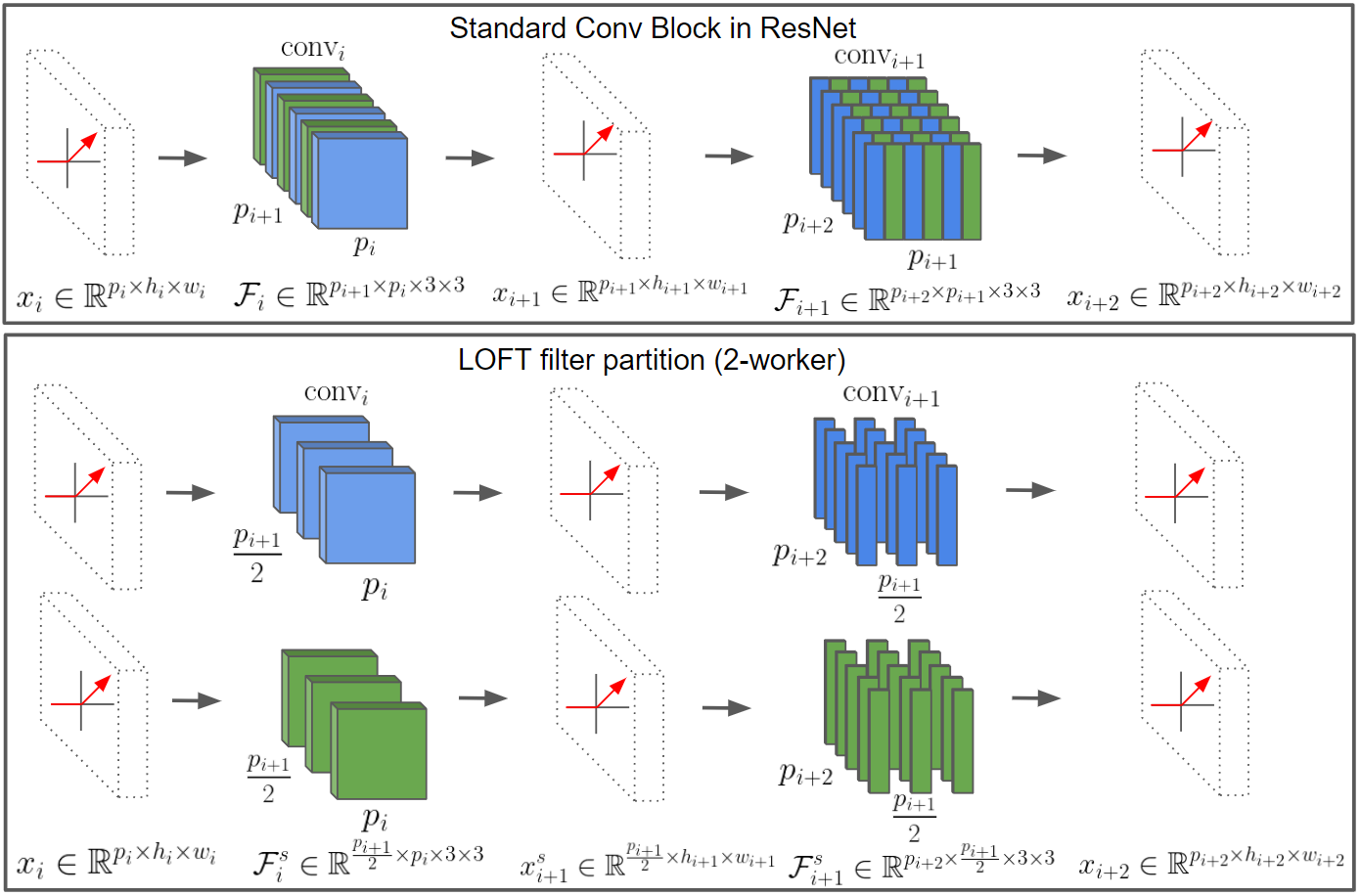}
    \caption{Depiction of the effect of a filter-wise \textsc{LoFT} partition on parameter sizes. Dotted blocks represent feature maps, colored blocks represent filters within convolutional layers.}
    \label{fig:filter_part}
    \vspace{-0.6cm}
\end{figure*}

This motivates our approach: we propose sampling and training  different sets of tickets during different stages of the pretraining. 
In this way, the algorithm is expected to ``touch'' upon the potential winning tickets at certain iterations. 
We conjecture (this is empirically shown in our experiments) that important filters in such winning tickets can be preserved and further recovered at the end of pretraining using our approach. 
These observations led us to the definition of the \textsc{LoFT} algorithm. 

\vspace{-0.2cm}
\section{The \textsc{LoFT} Algorithm}
% \vspace{-0.2cm}
% \subsection{Filter-wise Decomposition of CNNs}
\vspace{-0.3cm}
\begin{algorithm}[H]
\centering
\caption{\textsc{LoFT} Algorithm}\label{alg:loft}
\begin{algorithmic}[1]
    \STATE \textbf{Parameter}: $T$ synchronization iterations in pretraining, 
    % $T'$ training iterations after prunning for winning ticket, 
    $S$ workers, $\ell$ local iterations, $W$ CNN weights, 
    % $W'$ the winning ticket, $P$ prunning ratio.%
    \vspace{-0.2cm}
    \\\hrulefill
    \STATE $h(W)$ $\leftarrow$ randomly initialized CNN.
    \FOR{$t = 0, \dots, T-1$}
    \STATE$\left\{h_s(W_s)\right\}_{s = 1}^S = \textcolor{teal}{\text{\texttt{filterPartition}}}(h(W), ~S)$
    \STATE Distribute each $h_s(W_s)$ to a different worker.
    \FOR{$s = 1, \dots, S$}
    %\State $subnet$ = $subnets$[$s$]
    \STATE Train $h_s(W_s)$ for $\ell$ iterations using local SGD.
    \ENDFOR
    \STATE $h(W) = \textcolor{violet}{\texttt{aggregate}}\left(\left\{h_s(W_s)\right\}_{s = 1}^S\right)$.
\ENDFOR
% \STATE $W'=\texttt{prune}( W, P)$.
% \FOR{$t = 0, \dots, T'-1$}
 % \STATE Distribute $h(W')$ to all workers as $h_s(W_s)$.
 % \FOR{$s = 1, \dots, S$}
    %\State $subnet$ = $subnets$[$s$]
    % \STATE Train $h_s(W_s)$ for $1$ iterations using SGD.
    % \ENDFOR
% \STATE $h(W') = \texttt{average}\left(\left\{h_s(W')\right\}_{s = 1}^S\right)$.
 % \ENDFOR
\end{algorithmic}
\end{algorithm}
\vspace{-0.3cm}
We treat ``\emph{sampling and training sets of tickets}'' as a filter-wise decomposition of a given CNN, where each ticket is a subnetwork with a subset of filters. 
This is shown in Fig. \ref{fig:filter_part}. 
The \textsc{LoFT} algorithm that implements our ideas is shown in Algorithm \ref{alg:loft}.
% Although IST \cite{IST} is most commonly used for fully-connected layers, we generalize this methodology to training high-performing CNN models.
% In this direction, our distributed pretraining approach relies upon a curated, filter-wise decomposition of the global CNN.
Each block within a CNN typically consists of two identical convolutional layers, $\text{conv}_i$ and $\text{conv}_{i+1}$.
As shown in Figure \ref{fig:filter_part}, our methodology operates by partitioning the filters of these layers, $\mathcal{F}_i$ and $\mathcal{F}_{i+1}$, to different subnetworks --see \textcolor{teal}{\texttt{filterPartition()}} step in Algorithm \ref{alg:loft}-- in a structured, disjoint manner.
These subnetworks are trained independently --see local SGD steps in Algorithm \ref{alg:loft}-- before aggregating their updates into the global model by directly placing the filters back to their original place---see \textcolor{violet}{\texttt{aggreegate()}} step in Algorithm \ref{alg:loft}. %, as proposed in \cite{IST}.
\emph{The full CNN is never trained directly.}

The filter-wise partition strategy for a convolutional block begins by disjointly partitioning the filters $\mathcal{F}_i$ of the first convolutional layer $\text{conv}_{i}$.
This operation can be implemented by permuting and chunking the indices of filters within the first convolutional layer and within the block, as shown in Figure \ref{fig:filter_part}. 
Formally, we randomly and disjointly partition the total $p_{i+1}$ filters into $S$ subsets, where each subset forms $\mathcal{F}_i^s \in \mathbb{R}^{(p_{i+1}/S) \times p_i \times 3 \times 3}$. 
$S$ here indicates the number of independent workers in the distributed system.
$\mathcal{F}_i^s$ forms a new convolutional layer, which produces a new feature map $x_{i+1}^s \in \mathbb{R}^{(p_{i+1}/S) \times h_{i+1} \times w_{i+1}}$ with $S$ times fewer channels.

Based on which channels are presented in the feature map $x_{i+1}^s$, we further partition the input channels of filters $\mathcal{F}_{i+1}$ in the second convolutional layer into $S$ sets of sub-filters (Figure \ref{fig:filter_part}). 
Formally, each set has $p_{i+2}$ sub-filters with $\sfrac{p_{i+1}}{S}$ input channels, $\mathcal{F}_{i+1}^s \in \mathbb{R}^{p_{i+2} \times \sfrac{p_{i+1}}{S} \times 3 \times 3}$ and produces a new feature map $x_{i+2}^s \in \mathbb{R}^{p_{i+2} \times h_{i+2} \times w_{i+2}} $.
The input/output dimensions of the convolutional block are unchanged. We repeat the partition for all convolutional blocks in a CNN to get a set of subnetworks with disjoint filters. In each subnetwork, the intermediate dimensions of activations and filters are reduced, resembling a ``bottleneck" structure.

Our methodology of choosing tickets/subnetworks avoids partitioning layers that are known to be most sensitive to pruning, such as strided convolutional blocks  \cite{rethinking-pruning}.
Parameters not partitioned are shared among subnetworks, so their values must be averaged when the updates of tickets/subnetworks are aggregated into the global model.\footnote{We detail how \text{LoFT} is implemented to provide enough information for potential users; yet, we conjecture that our ideas could be applied to other architectures with appropriate modifications, showing the applicability to diverse scenarios.}

Compared with common distributed protocols, our pretraining methodology $i)$ reduces the communication costs, since we only communicate the tickets/subnetworks; and $ii)$ reduces the computational and memory costs on each worker, since we only locally train the sampled tickets/subnetworks that are smaller than the global model. 
From a different pespective, our approach allows pretraining networks beyond the capacity of a single-GPU:
The global model could be a factor of $O(S)$ wider than each subnetwork, allowing the global model size to be extended far beyond the capacity of single GPU. 
\emph{The ability to train such ``ultra-wide" models is quite promising for pruning purposes.}

% \vspace{-0.2cm}
% \subsection{}
% \vspace{-0.2cm}

\textbf{After pretraining with \textsc{LoFT}.} We perform standard pruning on the whole network
%---see \texttt{prune()} step in Algorithm \ref{alg:loft}---
to recover the winning ticket, and use standard training techniques over this winning ticket until the end of training.

\vspace{-0.2cm}
\section{Theoretical Result}
\vspace{-0.2cm}

\begin{figure}[!htp]
    \centering
    \includegraphics[width=0.9\linewidth]{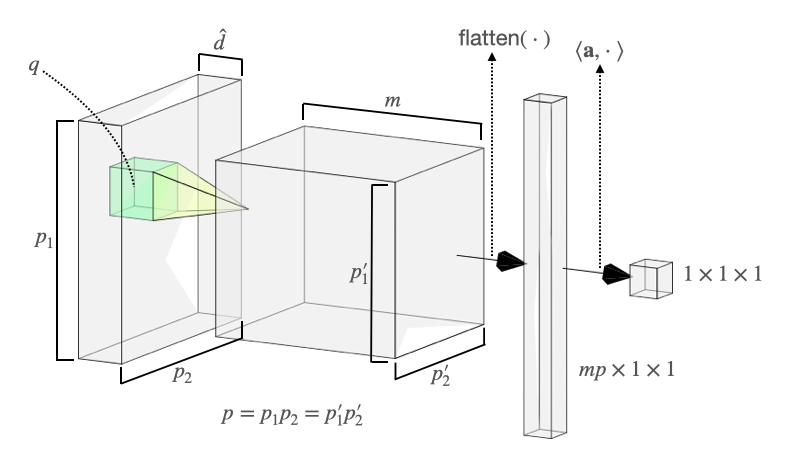}
    \caption{One-Hidden-Layer CNN for Theoretical Analysis}
    \label{fig:simple_cnn}
    \vspace{-0.4cm}
\end{figure}

We perform theoretical analysis on a one-hidden-layer CNN (see \figref{fig:simple_cnn}), and show that \textit{the trajectory of the neural network weight in \textsc{LoFT} stays near to the trajectory of gradient descent (GD)}. 
Since filter pruning is based on magnitude ranking of the filters, a small difference between the filters learned with \textsc{LoFT} and the filters learned with GD will more likely preserve the winning tickets. 

Consider a training dataset $(\X,\y) = \{(\x_i, y_i)\}_{i=1}^n$, where each $\x_i\in\R^{\hat{d}\times p}$ is an image and $y_i$ being its label. 
Here, $\hat{d}$ is the number of input channels and $p$ the number of pixels. 
Let $q$ denote the size of the filter, and let $m$ be the number of filters in the first layer. As in previous work \cite{du2018GDFindsGlobalMinima}, we let $\hat{\phi}(\cdot)$ denote the patching operator with $\hat{\phi}(x) \in\R^{q\hat{d}\times p}$. Consider the first layer weight $\W\in\R^{m\times q\hat{d}}$, and second layer (aggregation) weight  $\mathbf{a}\in\R^{m\times p}$. We assume that only the first layer weights $\W$ is trainable. In this case, the CNN trained on the means squared error has the form:
\begin{align*}
    f(\mathbf{x},\boldsymbol{\zeta}) = \inner{\mathbf{a}}{\sigma\paren{\W\hat{\phi}\paren{\x}}};\\ \mathcal{L}\paren{\W} = \norm{f(\X,\W) - \y}_2^2,
\end{align*}
where $\boldsymbol{\zeta}$ abstractly represents all training parameters, $f(\mathbf{x}, \cdot)$ denotes the output of the one-layer CNN for input $\mathbf{x}$, and $\mathcal{L}(\cdot)$ is the loss function.
We make the following assumption on the training data and the CNN weight initialization.
\begin{asump}
(Training Data)
\label{data_asump}
\textit{Assume that for all $i\in[n]$, we have $\norm{\x_i}_F = q^{-\frac{1}{2}}$ and $|y_i|\leq C$ for some constant $C$. Moreover, for all $i,i'\in[n]$ we have $\x_i\neq \x_{i'}$.} 
\end{asump}

Note that the first part of this assumption is standard and can be satisfied by normalizing the data \cite{du2018GDFindsGlobalMinima}. For simplicity of the analysis, let $d: = q\hat{d}$.

\begin{asump}
(Initialization)
\label{init_asump}
\textit{$\w_{0,r}\sim\mathcal{N}\paren{0, \kappa^2\mathbf{I}}$ and $a_{rj}\sim\left\{\pm\frac{1}{p\sqrt{m}}\right\}$ for $r\in[m]$ and $j\in[p]$.}
\end{asump}

We consider a simplified \textsc{LoFT} training scheme: assume that in the $t$th global iteration, we sample a set of $S$ masks, $\{\mathbf{m}_t^{(s)}\}_{s=1}^S$, for the filters, where each $\mathbf{m}_t^{(s)}\in \{0,1\}^m$. 
Let $m_{r,t}^{(s)}$ be the $r$th entry of $\mathbf{m}_t^{(s)}$. 
We assume that $m_{r,t}^{(s)}\sim\texttt{Bern}\paren{\xi}$ for some $\xi\in(0,1]$ for all $s\in[S]$ and $r\in[m]$, with $m_{r,t}^{(s)} = 1$ if the $r$th filter is trained in subnetwork $s$ and $m_{r,t}^{(s)} = 1$ otherwise. 
Intuitively, $\xi$ is the probability that a filter is selected to be trained in a subnetwork. 
Let $\mathbf{m}_t^{(s)}$ be the $s$th column of the joint mask matrix $\mathbf{M}_t$. 
Then, each row $\mathbf{m}_{r,t}$ contains information of the subnetwork indices in which the $r$th filter is active.
We further assume that the number of local iterations $\ell = 1$. 

Let $\{\W_t\}_{t=0}^T$ and $\{\hat{\W}_t\}_{t=0}^T$ be the weights in the trajectory of \textsc{LoFT} and GD, and let $\theta = \mathcal{P}\paren{\vee_{s=1}^S\{m_r^{(s)} =1\}} = 1 - (1-\xi)^S$. 
Next, we show that the expected difference between the two is bounded as follows:
\begin{theorem}
\label{main_theorem}
\textit{Let $f(\cdot, \cdot)$ be a one-hidden-layer CNN with the second layer weight fixed. 
Assume the number of hidden neurons satisfies
$m = \Omega\paren{\frac{n^4K^2}{\lambda_0^4\delta^2}\max\{n, d\}}$ and the step size satisfies $\eta = O\paren{\frac{\lambda_0}{n^2}}$:
Let Assumptions 1 and 2 be satisfied. 
Then, with probability at least $1-O\paren{\delta}$ we have:} \vspace{-0.4cm}
\begin{align*}
    & \textcolor{red}{\E_{[\mathbf{M}_T]}\left[\norm{\W_T - \hat{\W}_T}_F^2\right]} +\\ & \quad\eta \textcolor{olive}{\sum_{t=0}^{T-1}\E_{[\mathbf{M}_T]}\left[\norm{f\paren{\X,\W_t} - f\paren{\X,\hat{\W}_t}}_2^2\right]} \\
    & \leq O\paren{\textcolor{violet}{\tfrac{n^2\sqrt{d}}{\lambda_0^2\kappa m^{\frac{1}{4}}\sqrt{\delta}}} + \textcolor{teal}{\tfrac{2\eta^2T\theta^2(1-\xi)\lambda_0}{S}}}.
\end{align*}
\end{theorem}
\textbf{Remarks.} Intuitively, this theorem states that the sum of the expected weight difference in the $T$th iteration (i.e., {\small \textcolor{red}{$\E_{[\mathbf{M}_T]}[\|\W_T - \hat{\W}_T\|_F^2]$}}) and the aggregation of the step-wise difference of the neural network output between \textsc{LoFT} and GD (i.e., {\small \textcolor{olive}{$\sum_{t=0}^{T-1}\E_{[\mathbf{M}_T]}\|f\paren{\X,\W_t} - f\paren{\X,\hat{\W}_t}\|_2^2$}}) is bounded and controlled by the quantity on the right-hand side.
In words, both the weights found by \textsc{LoFT} as well as the output of \textsc{LoFT} are close to the ones found by regular training. 
Notice that increasing the number of filters $m$ (term in \textcolor{violet}{violet color}) and the number of subnetworks $S$ (term in \textcolor{teal}{teal color}) will drive the bound of the summation to zero. For a more thorough discussion, as well as the proof, please see supplementary material.
As a corollary of the theorem, we also show that \textsc{LoFT} training converges linearly upon to some neighborhood of the global minimum:
\begin{corollary}
\label{conv_coro}
\textit{Let the same condition of theorem (\ref{main_theorem}) holds, then we have:}
\begin{align*}
    & \E_{[\mathbf{M}_{t-1}]}\left[\norm{f(\X,\W_t) - \y}_2^2\right] \leq \\
    & \quad\paren{1 - \tfrac{\theta\eta\lambda_0}{2}}^t\norm{f(\X,\W_0) - \y}_2^2 + \\
    & \quad O\paren{\tfrac{\xi(1-\xi) n^3d}{m\lambda_0^2} + \tfrac{n\kappa^2\paren{\theta-\xi^2}}{S}}.
\end{align*}
\end{corollary} \vspace{-0.3cm}
We defer the proof for Theorem \ref{main_theorem} and Corollary \ref{conv_coro} to the appendix. 
%In particular, we provide a more detailed formulation of LoFT and prove an extended version of corollary \ref{conv_coro}), which will be used for the proof of theorem (\ref{main_theorem}).

\begin{table*}[!ht]
\vspace{0.0cm}
\begin{center}
\begin{sc}
\begin{scriptsize}
\begin{tabular}{lcllllrr}
\toprule
\multirow{2}{*}{setting}                                                          & \multirow{2}{*}{Dense Model} & \multirow{2}{*}{methods} & \multicolumn{3}{c}{pruning ratio} & \multirow{2}{*}{Comm. cost} & \multirow{2}{*}{improv.} \\ \cline{4-6}
                                                                                  &                             &                          & 80\%      & 50\%      & 30\%        &                            &                          \\
\toprule
\multirow{4}{*}{\begin{tabular}[c]{@{}l@{}}PreResNet-18\\ \textbf{CIFAR-10}\end{tabular}}  & \multirow{4}{*}{94.36}      & Gpipe-2                  & 94.41     & 94.55     &\multicolumn{1}{c}{}           & \textcolor{red}{131.88G}                   &   \textcolor{red}{$3.29\times$}                       \\
                                                                                  &                             & Local SGD-2              & 94.37     & 94.41     &\multicolumn{1}{c}{}           & \textcolor{orange}{55.40G}                   &    \textcolor{orange}{$1.38\times$}                   \\
                                                                                  &                             & LoFT-2                   & 93.97     & 94.11     &\multicolumn{1}{c}{}           & \textbf{\textcolor{teal}{40.02G}}             &        -              \\
                                                                                  \cline{3-8}
                                                                                  &                             & Gpipe-4                  & 94.41     & 94.55     &\multicolumn{1}{c}{}           & \textcolor{red}{659.42G}                   &    \textcolor{red}{$10.21\times$}                 \\
                                                                                  &                             & Local SGD-4              & 94.52     & 94.81     &\multicolumn{1}{c}{}           & \textcolor{orange}{110.80G}                   &    \textcolor{orange}{$1.72\times$}                     \\
                                                                                  &                             & LoFT-4                   & 93.97     & 94.13     &\multicolumn{1}{c}{}           & \textbf{\textcolor{teal}{64.57G}}                    &       -                   \\
\toprule
\multirow{4}{*}{\begin{tabular}[c]{@{}l@{}}PreResNet-34\\ \textbf{CIFAR-10}\end{tabular}}  & \multirow{4}{*}{93.51}      & Gpipe-2                  & 93.93     & 94.38     &\multicolumn{1}{c}{}           &  \textcolor{red}{131.88G}                  &    \textcolor{red}{$2.01\times$}                      \\
                                                                                  &                             & Local SGD-2              & 94.77     & 95.13     &\multicolumn{1}{c}{}           & \textcolor{orange}{105.93G}                   &      \textcolor{orange}{$1.61\times$}                   \\
                                                                                  &                             & LoFT-2                   & 93.25     & 93.43     &\multicolumn{1}{c}{}           & \textbf{\textcolor{teal}{65.36G}}                     & -                         \\
                                                                                  \cline{3-8}
                                                                                  &                             & Gpipe-4                  & 93.93     & 94.38     &\multicolumn{1}{c}{}           & \textcolor{red}{461.60G}                   &      \textcolor{red}{$5.12\times$}                    \\
                                                                                  &                             & Local SGD-4              & 94.64     & 94.82     &\multicolumn{1}{c}{}           & \textcolor{red}{211.86G}                  &      \textcolor{red}{$2.34\times$}                \\
                                                                                  &                             & LoFT-4                   & 93.89     & 94.02     &\multicolumn{1}{c}{}           & \textbf{\textcolor{teal}{90.17G}}                        &  -                        \\
\toprule
\multirow{4}{*}{\begin{tabular}[c]{@{}l@{}}ResNet-34\\ \textbf{CIFAR-10}\end{tabular}}     & \multirow{4}{*}{93.22}      & Gpipe-2                  & 93.69     & 93.81     &\multicolumn{1}{c}{}           &  \textcolor{red}{131.88G}                  &    \textcolor{red}{$2.01\times$}                          \\
                                                                                  &                             & Local SGD-2             & 94.49     & 94.74     &\multicolumn{1}{c}{}           & \textcolor{orange}{105.93G}                  &           \textcolor{orange}{$1.61\times$}               \\
                                                                                  &                             & LoFT-2                   & 93.38     & 93.41     &\multicolumn{1}{c}{}           & \textbf{\textcolor{teal}{65.36G}}                          & -                         \\ \cline{3-8}
                                                                                  &                             & Gpipe-4                  & 93.69     & 93.81     &\multicolumn{1}{c}{}           & \textcolor{red}{461.60G}                          & \textcolor{red}{$5.12\times$}                         \\
                                                                                   &                             & Local SGD-4             & 94.69     & 94.61     &\multicolumn{1}{c}{}           & \textcolor{red}{211.86G}                   &        \textcolor{red}{$2.34\times$}            \\
                                                                                  &                             & LoFT-4                   & 93.41     & 93.60     &\multicolumn{1}{c}{}           & \textbf{\textcolor{teal}{90.17G}}                   & -                         \\
\toprule
\multirow{4}{*}{\begin{tabular}[c]{@{}l@{}}PreResNet-18\\ \textbf{CIFAR-100}\end{tabular}} & \multirow{4}{*}{75.36}      & Gpipe-2                  & 75.38     & 75.91     &\multicolumn{1}{c}{}           & \textcolor{red}{131.88G}                   &     \textcolor{red}{$3.29\times$}                    \\
                                                                                  &                             & Local SGD-2              &  75.63    & 75.79     &\multicolumn{1}{c}{}           & \textcolor{orange}{55.51G}                  &      \textcolor{orange}{$1.38\times$}                  \\
                                                                                  &                             & LoFT-2                   & 75.99     & 76.65     &\multicolumn{1}{c}{}           & \textbf{\textcolor{teal}{40.03G}}                &        -                     \\ \cline{3-8}
                                                                                  &                             & Gpipe-4                  & 75.38     & 75.91     &\multicolumn{1}{c}{}           & \textcolor{red}{659.42G}                   &         \textcolor{red}{$10.21\times$}                 \\
                                                                                   &                             & Local SGD-4             & 75.50     & 75.44     &\multicolumn{1}{c}{}           & \textcolor{orange}{111.03G}                  &         \textcolor{orange}{$1.72\times$}                  \\
                                                                                  &                             & LoFT-4                   & 75.95     & 76.72     &\multicolumn{1}{c}{}           & \textbf{\textcolor{teal}{64.57G}}                   &       -                     \\
\toprule
\multirow{4}{*}{\begin{tabular}[c]{@{}l@{}}PreResNet-34\\ \textbf{CIFAR-100}\end{tabular}} & \multirow{4}{*}{76.57}      & Gpipe-2                  & 76.72     & 77.09     &\multicolumn{1}{c}{}           &  \textcolor{red}{131.88G}                          &    \textcolor{red}{$2.01\times$}                      \\
                                                                                  &                             & Local SGD-2              &  75.26    & 76.18     &\multicolumn{1}{c}{}           & \textcolor{orange}{106.05G}                   & \textcolor{orange}{$1.61\times$}\\
                                                                                  &                             & LoFT-2                   & 75.93     & 77.27     &\multicolumn{1}{c}{}           & \textbf{\textcolor{teal}{65.37G}}     & -                         \\ \cline{3-8}
                                                                                  &                             & Gpipe-4                  & 76.72     & 77.09     &\multicolumn{1}{c}{}           & \textcolor{red}{461.60G}                          &    \textcolor{red}{$5.12\times$}                      \\
                                                                                 &                             & Local SGD-4              & 76.62     &  75.79    &\multicolumn{1}{c}{}           & \textcolor{red}{212.10G}               &             \textcolor{red}{$2.34\times$}             \\
                                                                                  &                             & LoFT-4                   & 75.77     & 76.79     &\multicolumn{1}{c}{}           & \textbf{\textcolor{teal}{90.17G}}                          & -                         \\
\toprule
\multirow{4}{*}{\begin{tabular}[c]{@{}l@{}}ResNet34\\ \textbf{CIFAR-100}\end{tabular}}     & \multirow{4}{*}{75.93}      & Gpipe-2                  & 75.51     & 76.00     &\multicolumn{1}{c}{}           &  \textcolor{red}{131.88G}                          &    \textcolor{red}{$2.01\times$}                      \\
                                                                                  &                             & Local SGD-2             & 75.23     & 76.35     &\multicolumn{1}{c}{}           & \textcolor{orange}{106.05G}                 &              \textcolor{orange}{$1.61\times$}           \\
                                                                                  &                             & LoFT-2                   & 76.11     & 77.07     &\multicolumn{1}{c}{}           & \textbf{\textcolor{teal}{65.37G}}                           & -                         \\ \cline{3-8}
                                                                                  &                             & Gpipe-4                  & 75.51     & 76.00     &\multicolumn{1}{c}{}           & \textcolor{red}{461.60G}                           &       \textcolor{red}{$5.12\times$}                   \\
                                                                                  &                             & Local SGD-4             & 76.19     & 76.81     &\multicolumn{1}{c}{}           & \textcolor{red}{212.10G}                  &             \textcolor{red}{$2.34\times$}             \\
                                                                                  &                             & LoFT-4                   & 75.05     & 76.51     &\multicolumn{1}{c}{}           & \textbf{\textcolor{teal}{90.17G}}                           & -                         \\
\toprule
\multirow{4}{*}{\begin{tabular}[c]{@{}l@{}}PreResNet-18\\ \textbf{ImageNet} \end{tabular}}  & \multirow{4}{*}{70.71}      & Gpipe-2                  & 66.71     & 69.14     &\multicolumn{1}{c}{70.29}      & \textcolor{red}{20954.24G}                  &     \textcolor{red}{$81.80\times$}                     \\
                                                                                  &                             & LoFT-2                   & 65.41     & 69.12     &\multicolumn{1}{c}{69.64}      & \textbf{\textcolor{teal}{256.62G}}                    &                     \\  \cline{3-8}
                                                                                  &                             & Gpipe-4                  & 66.71     & 69.14     &\multicolumn{1}{c}{70.29}      & \textcolor{red}{52385.59G}                  &          \textcolor{red}{$126.27\times$}                \\
                                                                                  &                             & Local SGD-4             & 65.40     & 66.94     &\multicolumn{1}{c}{67.52}           & \textcolor{orange}{711.46G}                  &             \textcolor{orange}{$1.71\times$}             \\
                                                                                  &                             & LoFT-4                   & 65.60     & 68.93     &\multicolumn{1}{c}{69.77}      & \textbf{\textcolor{teal}{414.84G}}                   &                    \\
\toprule
\end{tabular}
\end{scriptsize}
\end{sc}
\end{center} \vspace{-0.2cm}
\caption{Left: Fine-tuned accuracy for different pretraining methods at different pruning ratios.  \textsc{Dense Model} corresponds to full CNN training without pruning. Right: Total communication costs (Comm.) of model parallel baseline (GPipe) \cite{huang2019gpipe}, Local SGD \cite{localsgdconverge} and \textsc{LoFT} during pretraining. Number after method name represents the number of parallel worker used. \textcolor{orange}{Orange color} indicates that the method in comparison consumes at most $2 \times$ communication bandwidth; \textcolor{red}{red color} indicates that the method in comparison consumes $> 2 \times$ communication bandwidth. Performance in \textcolor{teal}{teal color} represents the best in terms of communication efficiency.} \vspace{-0.2cm}
\label{tab:ist_acc}
\end{table*}

\vspace{-0.2cm}
\section{Experiments}
\vspace{-0.3cm}

% We verify that \textsc{LoFT} can preserve the winning tickets and non-trivially reduce costs during pretraining.
% Our experiments are split into three parts: we first show that \textsc{LoFT} recovers winning tickets for various pruning levels. 
% Key takeaway is that \textsc{LoFT} tickets could reach at least the same final accuracy as compared to full model training, before pruning. 
% Secondly, we empirically illustrate that \textsc{LoFT} does not recover the winning tickets by chance: \textsc{LoFT} preserves the winning filters and converges to overall model winning tickets throughout pretraining.
% Finally, we demonstrate our significant advantage in reducing the communication cost as compared to the standard distributed pretraining methods.

We show that \textsc{LoFT} can preserve the winning tickets and non-trivially reduce costs during pretraining.
First, we show that \textsc{LoFT} recovers winning tickets under various settings for all pruning levels with a significant reduction in communication cost compared to other model-parallel methods. Second, we illustrate that \textsc{LoFT} does not recover the winning tickets by chance: \textsc{LoFT} converges to winning tickets faster and provide better tickets for all pretraining length.

\textbf{Experimental Setup.}
We consider the workflow of pretraining for 20 epochs and fine-tuning for 90 epochs. We consider three CNNs: PreActResNet-18, PreActResNet-34 \cite{preactres}, and WideResNet-34 \cite{DBLP:journals/corr/ZagoruykoK16} to characterize our performance on models of different sizes and structures. 
We test these settings on the CIFAR-10, CIFAR-100, and ImageNet datasets. 
% We compare to conventional, full-model pretraining. We choose to base our experiment on WideResNet-18 (instead of e.g., AlexNet or VGG based architectures), since WideResNet has fewer parameters in the fully connected layers, making pruning such a network challenging \cite{li2017pruning}. 
% Further, WideResNet-18 network is a small ResNet model, increasing the challenges in pruning. 

For our baseline, we compare with a standard model-parallel algorithm Gpipe \cite{huang2019gpipe}, where we distribute layers of a network to different workers. 
We note that model-parallel algorithms are equivalent to training the whole model on a single large GPU. % they will have identical model checkpoints and final accuracy, which is what we report below.
However, since \textsc{LoFT} partitions the model based on the number of workers and utilizes independent training for each subnetwork, it is not equivalent to full model training, and the final performance will be different based on the number of workers we use. 
This is where our savings in communication cost come from, and why it is non-trivial for \textsc{LoFT} to even match the performance of other methods.
We also compare against the standard data parallel local SGD methodology \cite{localsgdconverge}. (Due to limitation of computing resource, we only experiment on 4-worker Local SGD on Imagenet dateset as the focus is on comparing with other model-parallel methods.)
% This justifies our choice to show results for both the 2- and 4-worker settings below.

We used two different pruning ratios: i.e., 50\%, and 80\% pruning ratios to profile performance under normal and over-pruning settings, respectively. 
For ImageNet, we additionally consider a pruning ratio 30\%, since networks are usually pruned less in this setting \cite{li2016pruning}. 
Here the pruning ratio $p\%$ represents that for each set of filters $\mathcal{F}_{i}$, we remove the bottom $p\%$ of the filters by its $\ell_2$-norm $\left\|\mathcal{F}_{i, j}\right\|_2$, as described above.
We do not prune the layers that are known to be most sensitive to pruning: this is skipping the first residual block and the strided convolutional blocks, according to \cite{rethinking-pruning,li2016pruning}.

There are methods with more specific pruning schedules, or different pruning ratios for different layers \cite{li2016pruning}. 
Here, \emph{we do not delve into layer-specific pruning or parameter tuning and only use one shared pruning ratio}. The focus is on the general quality of the winning ticket selected from \textsc{LoFT} with other model-parallel methods.

\noindent
\textbf{Implementation Details.}
We provide a PyTorch implementation of \textsc{LoFT} using the NCCL distributed communication package for training ResNet \cite{he2015deep} and WideResNet \cite{DBLP:journals/corr/ZagoruykoK16} architectures.
% Our implementation is centralized, meaning that a single process serves as a central parameter server. From this central process, the weights of the global model are maintained and partitioned to different worker processes (including itself) for independent training. 
Experiments are conducted on a node with 8 NVIDIA Tesla V100-PCIE-32G GPUs, a 24-core Intel(R) Xeon(R) Gold 5220R CPU \@ 2.20GHz, 1.5 TB of RAM.
%we highlight that IST can work as a cheap alternative to full model pretraining that allows efficient, parallel pretraining without negatively impacting final accuracy. In the following experiments, we establish that IST provide similar of better filter distance convergence speed, comparable final accuracy at all pruning levels.
% \usepackage{multirow}
% \usepackage{multirow}

\textbf{\textsc{LoFT} recovers winning tickets with lower communication cost.}
Table \ref{tab:ist_acc} shows the performance comparison for \textsc{LoFT}, Local SGD (data parallel) and Gpipe (model parallel) under various settings. 
We also include the performance of the \textsc{Dense Model}, where the network is trained as-is with the same setting without any pruning.
% In Table \ref{tab:ist_acc}, we consider the following scenario.
% For each of the datasets, we pretrain for 20 epochs, after which we select the winning ticket by pruning 50\% and 80\% of filters---this corresponds to \textsc{Baseline} (GPipe \cite{huang2019gpipe}) and \textsc{LoFT} scenarios in Table \ref{tab:ist_acc}. 
% % \textcolor{magenta}{Does it make sense to describe what is "pruning XX\% of filters" and how?} Added description above in experiment setup
% Then, we train the selected winning tickets for further 90 epochs using regular training. 
% For the full CNN training without pruning, we train the full model for 90 epochs---this corresponds to \textsc{No-pruning} in Table \ref{tab:ist_acc}.
% Some key points worth highlighting are: In the CIFAR-100 setting, \textsc{LoFT} finds slightly better winning tickets at all pruning levels considered, as compared to the baseline pretraining method. Furthermore, such a winning ticket reaches a better final accuracy, as compared to the full model training. 
We can see that across different model sizes, network structures, pruning ratios, and datasets, \textsc{LoFT} finds comparable or better tickets compared to other model-/data-parallel pretraining methods, while providing sizable savings in communication cost. 
% This might suggest \textsc{LoFT} adds some regularization effect. 
Note that \textsc{LoFT} partitions the model into smaller subnetworks; so it is non-trivial that e.g., the 4-worker case leads to the same final accuracy, as compared to the 2-worker or the full model cases.

While \textsc{LoFT} inherits the memory efficiency from model-parallel training methods, it further reduces  the communication cost from $1.38 \times$ up to $126.27 \times$, as shown in Table \ref{tab:ist_acc}. 
Similar behavior is observed in comparison to data-parallel training methids: the gains in communication overhead range from $1.38 \times$ to $2.34 \times$. We note though that in this case, for a sufficiently large model, it could be the case that the model does not fit in the workers' GPU RAM; in contrast, GPipe and \textsc{LoFT} allow efficient training of larger neural network models, by definition.
This overall improvement by \textsc{LoFT} is achieved by $i)$ changing the way of decomposing the network such that each worker can host an independent subnetwork and train locally without communication, which greatly reduces the communication frequency; and $ii)$ each worker only exchange the weight of the subnetwork after each round of local training instead of transmitting activation maps and gradients.

\begin{figure*}[!h]
    \begin{subfigure}[b]{0.32\textwidth}
         \centering
         \includegraphics[width=\textwidth]{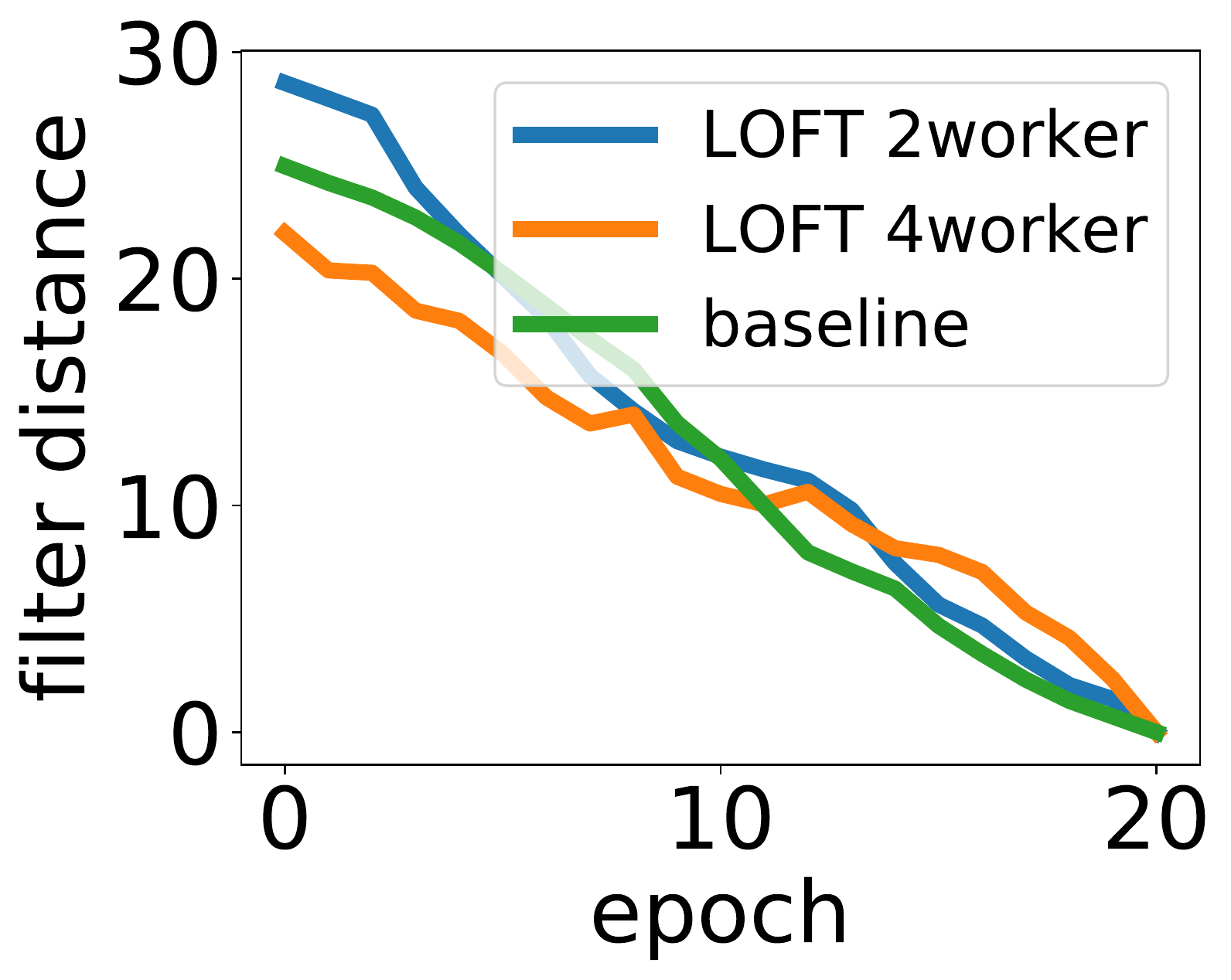}
         \caption{CIFAR10}
     \end{subfigure}
    \begin{subfigure}[b]{0.32\textwidth}
         \centering
         \includegraphics[width=\textwidth]{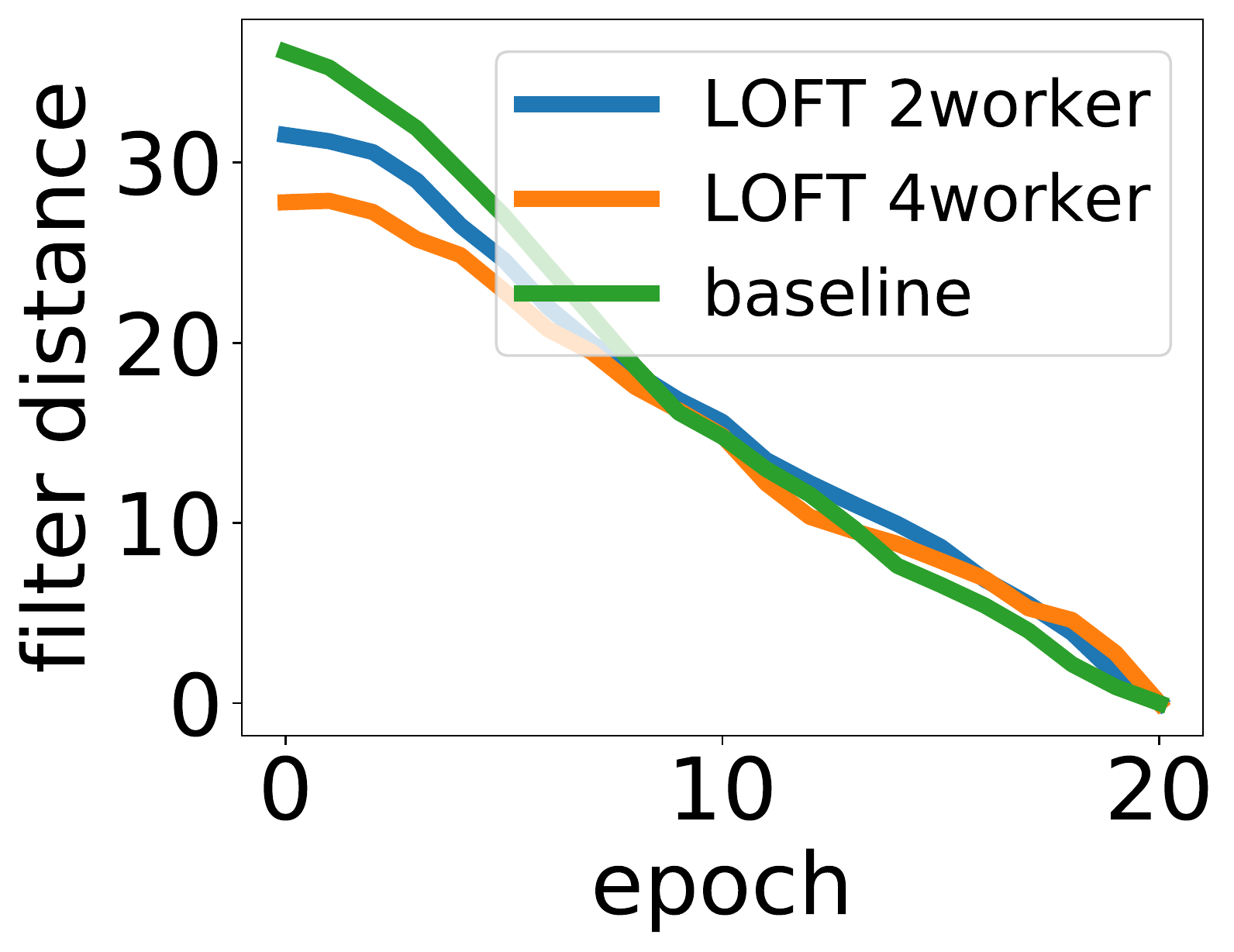}
         \caption{CIFAR100}
     \end{subfigure}
    %  \hfill
     \begin{subfigure}[b]{0.32\textwidth}
         \centering
         \includegraphics[width=\textwidth]{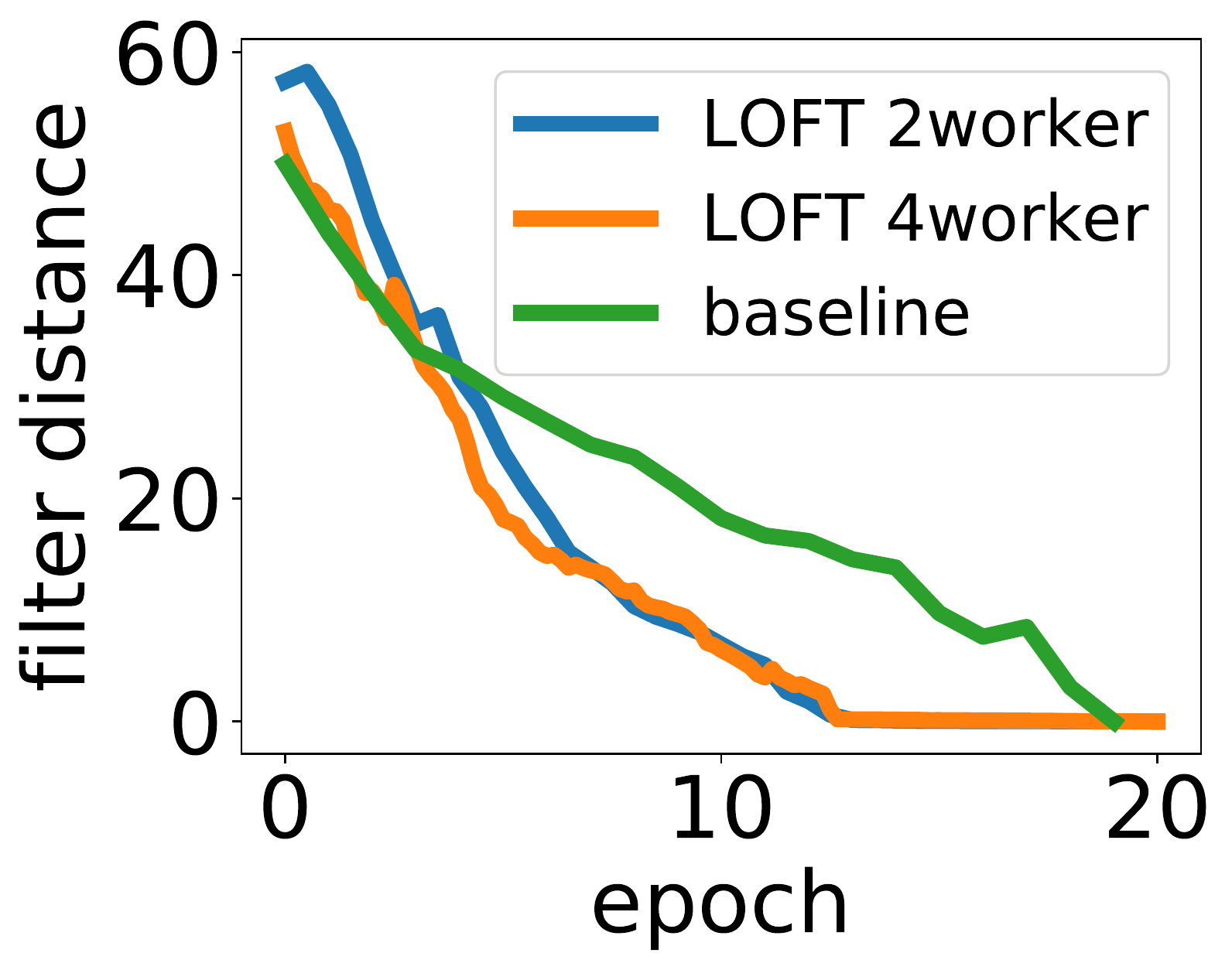}
         \caption{ImageNet}
     \end{subfigure}
    \caption{filter distance with respect to final ticket throughout pretraining process}
    \label{fig:filter_dist}
    \vspace{-0.5cm}
\end{figure*}

\textbf{\textsc{LoFT} converges faster provides better tickets throughout pretraining.}
\begin{figure}[!t]
  \begin{center}
    \includegraphics[width=1\linewidth]{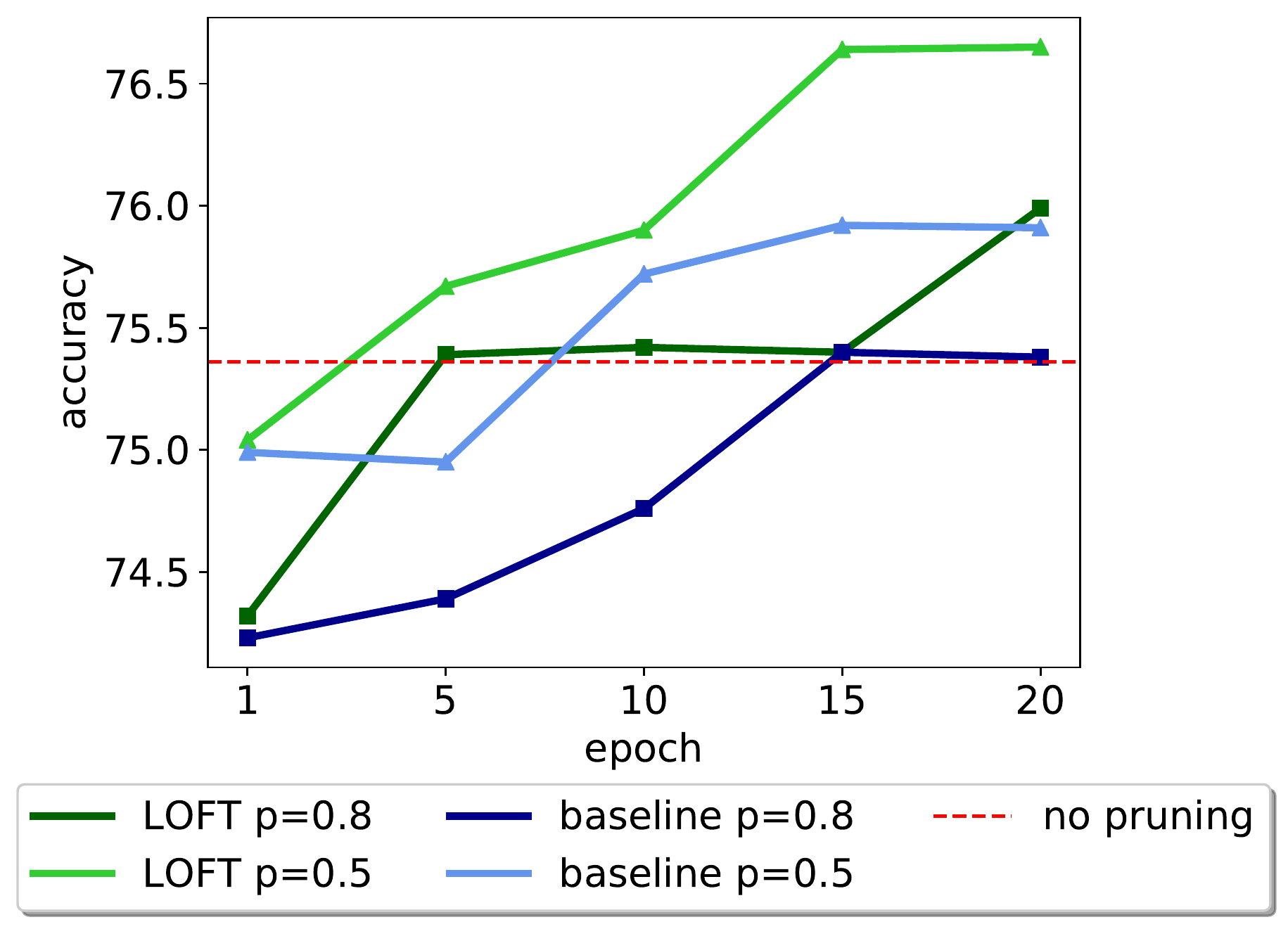}
  \end{center} \vspace{-0.3cm}
  \caption{Final accuracy after $i)$ selecting a ticket drawn at different epochs for training (x-axis), and $ii)$ further training the selected ticket for 90 epochs, as described in main text. This plot considers the CIFAR100 case.
    % \textcolor{magenta}{I assume having such plots for the other datasets is difficult?} Yes.
    }
    \label{fig:final_acc}
    \vspace{-0.4cm}
\end{figure}
We provide a closer look into some empirical results that showcase \textsc{LoFT}'s ability to provide better tickets early in training.
In Figure \ref{fig:final_acc}, we sampled different tickets (every 5 epochs) from the first 20 epochs in pretraining, and report all their final accuracy after fine tuning. We can see that \textsc{LoFT} yields better tickets (higher final accuracy) compared to Gpipe baseline. This shows that \textsc{LoFT} does not rely on a particular pretraining length and provides better tickets throughout pretraining.

To better compare the filter convergence, we calculated the filter distance between filters in the final epoch and filters in the previous pretraining epochs, As described above, the filter distance measures the change in the rankings of the filters. 
Larger filter distance means the important filters have not yet been identified as the top rank filters. 
As shown in Figure \ref{fig:filter_dist}. 
% \textsc{LoFT} identifies the important filters very quickly and preserves these important filters in the winning tickets. 
\textsc{LoFT} quickly and monotonically decreases the filter distance to the filters in the final epoch, showing it is able to efficiently identify and preserve the correct winning ticket. 
In the more challenging ImageNet dataset, \textsc{LoFT} can decrease the filter distance faster than the baseline pretraining, which suggests \textsc{LoFT} is a better way to find the winning ticket by providing some acceleration in filter convergence.

\vspace{-0.2cm}
\section{Related Work}
\vspace{-0.2cm}
% general LTH stuff

Algorithms for distributed training may be categorized into \emph{model parallel} and \emph{data parallel} methodologies. 
In the former \cite{dean2012large,hadjis2016omnivore}, portions of the NN are partitioned across different compute nodes, while, in the latter \cite{farber1997parallel,raina2009large}, the complete NN is updated with different data on each compute node.
Due to its ease-of-implementation, data parallel training is the most popular distributed training framework in practice. % \cite{abadi2016tensorflow,paszke2017automatic}.

As data parallelism needs to update the whole model on each worker --which still results in a large memory and computational cost-- researchers utilize model parallelism, such as Gpipe \cite{huang2019gpipe}, to reduce the per node computational burden. 
On the the other hand, pure model parallelism needs to synchronize at every training iteration to exchange intermediate activations and gradient information between workers, resulting in high communication costs. 

Following recent work that efficiently discovers winning tickets early in the training process \cite{early_bird}, our methodology further improves the efficiency of LTH by extending its application to communication-efficient, distributed training.
Furthermore, by allowing significantly larger networks during pretraining, we enable the discovery of higher-performing winning tickets.

Previous work by \cite{early_bird} proposed mask distance as a tool for identifying winning ticket early in the training process. Mask distance considers the Hamming distance of the 0-1 pruned mask on batch normalization (BN) layers. While similar in motivation, our filter distance criteria is fundamentally different. The two methods aim to capture totally different part of the training dynamic. Filter distance profiles the ordering of convolutional filters, while mask distance captures the activation of batch normalization layers. Filter distance models pruning process as a ranking whereas mask distance models it as a binary mask. Furthermore, filter distance is a consistent measurement that does not depend on the pruning ratio whereas mask distance can only be calculated with respect to a specific pruning ratio. 

\vspace{-0.2cm}
\section{Conclusion and Discussion}
\vspace{-0.2cm}
\textsc{LoFT} is a novel model-parallel pretraining algorithm that is both memory and communication efficient. Moreover, experiments show that \textsc{LoFT} can discover tickets faster or comparable than model-parallel training, and discover tickets with higher or comparable final accuracy. An immediate future work is, with more computation budget, testing \textsc{LoFT} with larger models and more challenging datasets. We are also curious how will the accuracy scale as we use more workers. 
Finally, it is also an open question whether we can further automate the pretraining process by using some adaptive stopping criteria to stop pretraining to identify winning tickets without hyperparameter tuning.

%\clearpage
\bibliographystyle{apalike}
\bibliography{example_paper}

\clearpage

\onecolumn
\appendix
\section{Detailed Mathematical Formulation of LoFT}
\zlabel{detailed_math_form}
For a vector $\mathbf{v}$, $\norm{\mathbf{v}}_2$ denotes its Euclidean ($\ell_2$) norm. For a matrix $\mathbf{V}$, $\norm{\mathbf{V}}_F$ denotes its Frobenius norm. We use $\mathcal{P}\paren{\cdot}$ to denote the probability of an event, and $\indy{\cdot}$ to denote the indicator function. For two vectors $\mathbf{v}_1,\mathbf{v}_2$, we use the simplified notation $\indy{\mathbf{v}_1;\mathbf{v}_2} := \indy{\inner{\mathbf{v}_1}{\mathbf{v}_2}\geq 0}$. Given that the mask in iteration $t$ is $\mathbf{M}_t$, we denote $\E_{[\mathbf{M}_t]}\left[\cdot\right] = \E_{\mathbf{M}_0,\dots,\mathbf{M}_t}\left[\cdot\right]$.

Recall that the CNN considered in this paper has the form
\begin{align*}
    f(\mathbf{x},\theta) = \inner{\mathbf{a}}{\sigma\paren{\W_1\hat{\phi}\paren{\x}}}
\end{align*}
Denote $\hat{\x} = \hat{\phi}\paren{\x}$. Essentially, this patching operator applies to each channel, with the effect of extending each pixel to a set of pixels around it. So we denote $\hat{\x}_i^{(j)}\in \R^{q\hat{d}}$ as the extended $j$th pixel across all channels in the $i$th sample. For each transformed sample, we have that $\norm{\hat{\x}_i}_F\leq \sqrt{q}\norm{\x_i}_F$. We simplify the CNN output as
\begin{align*}
    f(\hat{\x},\W) = \inner{\mathbf{a}}{\sigma\paren{\W\otimes\hat{\x}}} = \sum_{r=1}^{m_1}\sum_{j=1}^pa_{rj}\sigma\paren{\inner{\hat{\x}^{(j)}}{\w_r}}
\end{align*}
In this way, the formulation of CNN reduces to MLP despite a different form of input data $\hat{\x}$ and an additional dimension of aggregation in the second layer.
We consider train neural network $f$ on the mean squared error (MSE)
\begin{align*}
    \mathcal{L}\paren{\W} = \norm{f\paren{\hat{\X},\W} - \y}_2^2 = \sum_{i=1}^n\paren{f\paren{\hat{\x}_i,\W} - y_i}^2
\end{align*}
Now, we consider an $S$-worker \textsc{LoFT} scheme. The subnetwork by filter-wise partition is given by
\begin{align*}
    f_{\mathbf{m}^{(s)}}\paren{\hat{\x},\W} = \sum_{r=1}^{m_1}\sum_{j=1}^pm^{(s)}_ra_{rj}\sigma\paren{\inner{\hat{\x}^{(j)}}{\w_r}}
\end{align*}
Trained on the regression loss, the surrogate gradient is given by
\begin{align*}
    \nabla_{\w_r}\mathcal{L}_{\mathbf{m}^{(s)}}\paren{\W} = m^{(s)}_r \sum_{i=1}^n\sum_{j=1}^p\paren{f_{\mathbf{m}^{(s)}}\paren{\hat{\x}_i,\W} - y_i}a_{rj}\hat{\x}_{i}^{(j)}\indy{\hat{\x}_i^{(j)};\w_r}
\end{align*}
We correspondingly scale the whole network function
\begin{align*}
    f\paren{\hat{\x},\W} = \xi\sum_{r=1}^m\sum_{j=1}^pa_{rj}\sigma\paren{\inner{\hat{\x}_i^{(j)}}{\w_r}}
\end{align*}
Assuming it is also training on the MSE, we write out its gradient as
\begin{align*}
    \nabla_{\w_r}\mathcal{L}\paren{\W} = \xi\sum_{i=1}^n\sum_{j=1}^n\paren{f\paren{\hat{\x}_i,\W} - y_i}a_{rj}\hat{\x}_i^{(j)}\indy{\hat{\x}_i^{(j)};\w_r}
\end{align*}
In this work, we consider the one-step \textsc{LoFT} training, given by
\begin{align*}
    \w_{r,t+1} = \w_{r,t} - \eta\frac{N_{r,t}^\perp}{N_{r,t}}\sum_{s=1}^S\nabla_{\w_r}\mathcal{L}_{\mathbf{m}_t^{(s)}}\paren{\W_{r,t}}
\end{align*}
Here, let $N_{r,t} = \max\left\{\sum_{s=1}^Sm_{r,t}^{(s)},1\right\}$, and $N_{r,t}^\perp = \min\left\{\sum_{s=1}^Sm_{r,t}^{(s)},1\right\}$. Intuitively, $N_{r,t}$ denote the "normalizer" that we will divide the sum of the gradients from all subnetworks with, and $N_{r,t}^\perp$ denote the indicator of whether filter $r$ is trained in at least one subnetwork. Let $\theta = \mathcal{P}\paren{N_{r,t}^\perp = 1} = 1 - (1 - \xi)^p$, denoting the probability that at least one of $\left\{m_{r,t}^{(s)}\right\}_{s=1}^S$ is one. Denote $u_t^{(i)} = f\paren{\hat{\x}_i,\W_t}$. For further convenience of our analysis, we define
\begin{align*}
    \tilde{u}_{r,t}^{(i)} = \frac{N_{r,t}^\perp}{N_{r,t}}\sum_{s=1}^Sm_{r,t}^{(s)}\hat{u}_t^{(s, i)};\quad \mathbf{g}_{r,t} = \frac{N_{r,t}^\perp}{N_{r,t}}\sum_{s=1}^S\nabla_{\w_r}\mathcal{L}_{\mathbf{m}_t^{(s)}}\paren{\W_t}
\end{align*}
Then the \textsc{LoFT} training has the form
\begin{align*}
    \w_{r,t+1} = \w_{r,t} - \eta\mathbf{g}_{r,t};\quad
    \mathbf{g}_{r,t} = \sum_{i=1}^n\sum_{j=1}^pa_{rj}\paren{\tilde{u}_{r,t}^{(i)} - N_{r,t}^\perp y_i}\hat{\x}_i^{(j)}\indy{\hat{\x}_i^{(j)};\w_{r,t}}
\end{align*}
Suppose that assumptions in main text hold. Then for all $i\in[n]$ we have $\norm{\x_i}_F = q^{\frac{1}{2}}$, and for all $i,i'\in[n]$ such that $i\neq i'$, we have $\x_i\neq \x_{i'}$.  As in previous work \citep{du2018GDFindsGlobalMinima}, we have $\norm{\hat{\x}_i}_F  \leq \sqrt{q}\norm{\x_i}_F \leq 1$. Thus, for all $j\in[p]$ we have $\norm{\hat{\x}_i^{(j)}}\leq 1$. Moreover, since $\x_i\neq \x_{i'}$ for $i\neq i'$, we then have $\hat{\x}_i\neq \hat{\x}_{i'}$, which implies that $\hat{\x}_i^{(j)}\neq \hat{\x}_{i'}^{(j)}$ for all $i\neq i'$ and $j\in[p]$.
\section{One-Step \textsc{LoFT} Convergence}
\label{convergence_proof}
In this section, our goal is to prove and extended version of Corollary 1 in the main text, as a new theorem. We first state the extended version here, and proceed to prove it.
\begin{theorem}
Let $f$ be a one-hidden-layer CNN with the second layer weight fixed. Assume the number of hidden neurons satisifes
$m = \Omega\paren{\frac{n^3K^2}{\lambda_0^4\delta^2\kappa^2}\max\{n, d\}}$ and the step size satisfies $\eta = O\paren{\frac{\lambda_0}{n^2}}$. Let Assumption 1 and 2 be satisfied. Then with probability at least $1-O\paren{\delta}$ we have that
\begin{align*}
    \E_{[\mathbf{M}_{t-1}]}\left[\norm{\U_{t} - \y}_2^2\right] & \leq \paren{1 - \frac{\theta\eta\lambda_0}{2}}^t\norm{\U_0 - \y}_2^2 + O\paren{\frac{\xi^2(1-\xi)^2 n^3d}{m\lambda_0^2} + \frac{n\kappa^2\paren{\theta-\xi^2}}{S}}
\end{align*}
and the weight perturbation is bounded by
\begin{align*}
    \E_{[\mathbf{M}_{t-1}]}\left[\norm{\w_{r,t}-\w_{r,0}}_2\right] \leq O\paren{\lambda_0^{-1}\sqrt{\frac{n}{m}}}\norm{\U_0 - \y}_2 + \\
    \eta\theta  T\cdot O\paren{\frac{\xi(1-\xi)n^2\sqrt{d}}{m\lambda_0} + n\kappa\sqrt{\frac{\theta-\xi^2}{mS}}}\\
\end{align*}
\end{theorem}
\label{LoFT_Proof}
This is essentially a multi-sample drop-out proof for one-hidden-layer MLP, with an additional summation over the pixels $j\in[p]$. For completeness we present the proof here. We care about the MSE computed on the scaled full network
\begin{align*}
    u_k^{(i)} = \xi\sum_{r=1}^m\sum_{j=1}^pa_{rj}\sigma\paren{\inner{\hat{\x}_i^{(j)}}{\w_{r,t}}};\quad\mathcal{L}\paren{\W_t} = \norm{\U_t - \y}_2^2
\end{align*}
Performing gradient descent on this scaled full network involves computing
\begin{align*}
    \nabla_{\w_r}\mathcal{L}\paren{\W_t} = \xi\sum_{i=1}^n\sum_{j=1}^p\paren{u_t^{(i)} - y_i}a_{rj}\xij\indy{\xij;\w_{r,t}}
\end{align*}
\subsection{Change of Activation Pattern}
Let $R$ be some fixed scale. For analysis convenience, we denote 
\begin{align*}
    A_{ir}^{(j)} = \left\{\exists\w: \norm{\w - \w_{r,0}}_2\leq R; \indy{\hat{\x}_i^{(j)};\w}\neq\indy{\hat{\x}_i^{(j)};\w_{r,0}}\right\}
\end{align*}
Note that $A_{ir}^{(j)}$ happens if and only if $\left|\inner{\hat{\x}_i^{(j)}}{\w_{r,0}}\right| < R$. Therefore $\mathbb{P}\paren{A_{ir}^{(j)}} < \frac{2R}{\kappa\sqrt{2\pi}}$. Denote 
\begin{align*}
    P_{ij} = \left\{r\in[m]:\neg A_{ir}^{(j)}\right\};\quad P_{ij}^\perp = [m]\setminus P_{ij}
\end{align*}
The next lemma shows the magnitude of $P_{ij}^\perp$
\begin{lemma}
Let $m = \Omega\paren{R^{-1}\log\frac{np}{\delta}}$. Then with probability at least $1-O(\delta)$ it holds for all $i\in[n]$ and $j\in[p]$ that
\begin{align*}
    \left|P_{ij}^\perp\right| \leq 3m\kappa^{-1}R
\end{align*}
\begin{proof}
The magnitude of $P_{ij}^\perp$ satisfies
\begin{align*}
    \left|P_{ij}^\perp\right| = \sum_{r=1}^m\indy{A_{ir}^{(j)}}
\end{align*}
The indicator function $\indy{A_{ir}^{(j)}}$ has bounded first and second moment
\begin{align*}
    \E_{\W}\left[\indy{A_{ir}^{(j)}}\right] = \mathbb{P}\paren{A_{ir}^{(j)}} & \leq \frac{2R}{\kappa\sqrt{2\pi}}\\ \E_{\W}\left[\paren{\indy{A_{ir}^{(j)}-\E_{\W}\left[\indy{A_{ir}^{(j)}}\right]}}^2\right]& \leq \E_{\W}\left[\indy{A_{ir}^{(j)}}^2\right] \leq \frac{2R}{\kappa\sqrt{2\pi}}
\end{align*}
This allows us to apply the Berstein Inequality to get that
\begin{align*}
    \mathbb{P}\paren{\sum_{r=1}^m\indy{A_{ir}^{(j)}} > \frac{2mR}{\kappa\sqrt{2\pi}} + mt} < \exp\paren{-\frac{m\kappa t^2\sqrt{2\pi}}{8\paren{1 + \frac{t}{3}}R}}
\end{align*}
Therefore, with probability at least $1 - np\exp\paren{-m\kappa^{-1} R}$ it holds for all $i\in[n]$ and $j\in[p]$ that
\begin{align*}
    \left|P_{ij}^\perp\right| = \sum_{r=1}^m\indy{A_{ir}^{(j)}} \leq 3m\kappa^{-1}R
\end{align*}
Letting $m = \Omega\paren{R^{-1}\log\frac{np}{\delta}}$ gives that the success probability is at least $1 - O(\delta)$.
\end{proof}
\end{lemma}
\subsection{Initialization Scale}Let $\w_{0,r}\sim\mathcal{N}\paren{0, \kappa^2\mathbf{I}}$ and $a_{j}\sim\left\{-\frac{1}{p\sqrt{m}},\frac{1}{p\sqrt{m}}\right\}$ for all $r\in[m]$ and $j\in[p]$. We cite some results from prior work to deal with the initialization scale
\begin{lemma}
\label{init_mat_bound}
Suppose $\kappa \leq 1, R\leq \kappa\sqrt{\frac{d}{32}}$. With probability at least $1 - e^{md/32}$ we have that
\begin{align*}
    \|W_0\|_F \leq  \kappa\sqrt{2md} - \sqrt{m}R\\
\end{align*}
\end{lemma}

\begin{lemma}
\label{init_inner_prod_bound}
Assume $\kappa \leq 1$ and $R \leq \frac{\kappa}{\sqrt{2}}$. With probability at least $1 - ne^{-\frac{m}{32}}$ over initialization, it holds for all $i\in[n]$ that
\begin{align*}
    \sum_{r=1}^m\inner{\mathbf{w}_{0,r}}{\mathbf{x}_i}^2 \leq 2m\kappa^2 - mR^2\\
    \sum_{i=1}^n\sum_{r=1}^m\inner{\mathbf{w}_{0,r}}{\mathbf{x}_i}^2 \leq 2mn\kappa^2 - mnR^2
\end{align*}
\end{lemma}
Moreover, we can bound the initial MSE
\begin{lemma}
\label{initial_scale}
Assume that for all $i\in[n]$, $y_i$ satisfies $|y_i|\leq C$ for some $C > 0$. Then, we have
\begin{align*}
    \E_{\mathbf{W}_0,\hat{\mathbf{a}}}\left[\|\y - \U_0\|_2^2\right] \leq \paren{p^{-1} + C^2}n
\end{align*}
\begin{proof}
It is obvious that $\E_{\mathbf{W}_0,\hat{\mathbf{a}}}\left[u_0^{(i)}\right] = 0$ for all $i\in[n]$. Moreover,
\begin{align*}
    \E_{\mathbf{W}_0,\hat{\mathbf{a}}}\left[u_0^{(i)2}\right] & = \sum_{r,r'=1}^m\sum_{j,j'=1}^p\E_{\hat{\mathbf{a}}}\left[a_{rj}a_{r'j'}\right]\E_{\W_0}\left[\sigma\paren{\inner{\hat{\x}_{i}^{(j)}}{\w_{0,r}}}\sigma\paren{\inner{\hat{\x}_{i}^{(j')}}{\w_{0,r}}}\right]\\
    & = \frac{1}{p^2m}\sum_{r=1}^m\sum_{j=1}^p\E_{\W_0}\left[\sigma\paren{\inner{\hat{\x}_{i}^{(j)}}{\w_{0,r}}}^2\right]\\
    & \leq \frac{1}{p^2m}\sum_{r=1}^m\sum_{j=1}^p\E_{\W_0}\left[\inner{\hat{\x}_{i}^{(j)}}{\w_{0,r}}\right]\\
    & \leq p^{-1}
\end{align*}
Therefore
\begin{align*}
    \E_{\mathbf{W}_0,\hat{\mathbf{a}}}\left[\norm{\U_0 -\y}_2^2\right] = \sum_{i=1}^n\E_{\mathbf{W}_0,\hat{\mathbf{a}}}\left[\paren{u_0^{(i)} -y_i}_2^2\right] = \sum_{i=1}^n\paren{\E_{\mathbf{W}_0,\hat{\mathbf{a}}}\left[u_0^{(i)2}\right] + y_i^2} \leq\paren{p^{-1}+ C^2}n
\end{align*}
\end{proof}
\end{lemma}
\subsection{Kernel Analysis}
The neural tangent kernel is defined to be the inner product of the gradient with respect to the neural network output. We let the finite-width NTK be defined as
\begin{align*}
    \h(t)_{ii'} = \sum_{r=1}^m\sum_{j,j'=1}^pa_{rj}a_{rj'}\inner{\hat{\x}_i^{(j)}}{\hat{\x}_{i'}^{(j')}}\indy{\hat{\x}_i^{(j)};\w_{r,t}}\indy{\hat{\x}_{i'}^{(j')};\w_{r,t}}
\end{align*}
Moreover, let the infinite width NTK be defined as
\begin{align*}
    \h^\infty_{ii'} =  \frac{1}{p^2}\sum_{j=1}^p\inner{\hat{\x}_i^{(j)}}{\hat{\x}_{i'}^{(j)}}\E_{\mathbf{w}\sim\mathcal{N}\paren{0,\mathbf{I}}}\left[\indy{\hat{\x}_i^{(j)};\w}\indy{\hat{\x}_{i'}^{(j)};\w}\right]
\end{align*}
Let $\lambda_0 = \lambda_{\min}\paren{\h^\infty}$. Note that since $\hat{\x}_i\not\parallel\hat{\x}_{i'}$ for $i\neq i'$. Thus $\hat{\x}_i^{(j)}\not\parallel\hat{\x}_{i'}^{(j)}$ for $i\neq i'$. () shows that the matrix $\hat{\h}(j)$ , as defined below, is positive definite for all $j\in[p]$
\begin{align*}
    \hat{\h}(j)_{ii'}^\infty = \inner{\hat{\x}_i^{(j)}}{\hat{\x}_{i'}^{(j)}}\E_{\mathbf{w}\sim\mathcal{N}\paren{0,\mathbf{I}}}\left[\indy{\hat{\x}_i^{(j)};\w}\indy{\hat{\x}_{i'}^{(j)};\w}\right]
\end{align*}
Since $\h^\infty = p^{-2}\sum_{j=1}^p\hat{\h}(j)^\infty$, we have that $\h^\infty$ is positive definite and thus $\lambda_0 > 0$. The following lemma shows that the NTK remains positive definite throughout training.
\begin{lemma}
Let $m = \Omega\paren{\lambda_0^{-2}n^2\log\frac{n}{\delta}}$. If for all $r\in[m]$ and all $t$ we have $\norm{\w_{r,t} - \w_{r,0}}_2\leq R := O\paren{\frac{\kappa\lambda_0}{n}}$. Then with probability at least $1-\delta$ we have that for all $t$
\begin{align*}
    \lambda_{\min}\paren{\h(t)}\geq \frac{\lambda_0}{2}
\end{align*}
\begin{proof}
To start, we notice that for all $r\in[m]$
\begin{align*}
    \E_{\mathbf{W}_0,\mathbf{a}}&\left[\sum_{j,j'=1}^pa_{rj}a_{rj'}\indy{\hat{\x}_i^{(j)};\w_{r,0}}\indy{\hat{\x}_{i'}^{(j')};\w_{r,0}}\right] \\
    &\quad\quad\quad= \frac{1}{p^2m}\E_{\w\sim\mathcal{N}(0,\mathbf{I})}\left[\indy{\hat{\x}_i^{(j)};\w}\indy{\hat{\x}_{i'}^{(j)};\w}\right]
\end{align*}
Moreover, we have that
\begin{align*}
    \left|\sum_{j,j'=1}^pa_{rj}a_{rj'}\inner{\hat{\x}_i^{(j)}}{\hat{\x}_{i'}^{(j')}}\indy{\hat{\x}_i^{(j)};\w_{r,0}}\indy{\hat{\x}_{i'}^{(j')};\w_{r,0}} \right|\leq 1
\end{align*}
Thus, we can apply Hoeffding's inequality with bounded random variable to get that
\begin{align*}
    \mathbb{P}\paren{\left|\h(0)_{i,i'} - \h^\infty_{i,i'}\right| \geq t}\leq 2\exp\paren{-mt^2}
\end{align*}
Therefore, with probability at least $1-O(\delta)$ it holds that for all $i,i'\in[n]$
\begin{align*}
    \left|\h(0)_{i,i'} - \h^\infty_{i,i'}\right|\leq \frac{\log\frac{n}{\delta}}{\sqrt{m}}
\end{align*}
which implies that
\begin{align*}
    \norm{\h(0) - \h^\infty}\leq \norm{\h(0) - \h^\infty}_F \leq \frac{n\paren{\log\delta^{-1}+\log n}}{\sqrt{m}}
\end{align*}
As long as $m = \Omega\paren{\lambda_0^{-2}n^2\log\frac{n}{\delta}}$ we will have
\begin{align*}
    \norm{\h(t) - \h^\infty}\leq \frac{\lambda_0}{4}
\end{align*}
Now we move on to bound $\norm{\h(t) - \h(0)}$. We have that
\begin{align*}
    \h(t)_{i,i'} - \h(0)_{i,i'} = \sum_{r=1}^m\sum_{j,j'=1}^pa_{rj}a_{rj'}\inner{\hat{\x}_i^{(j)}}{\hat{\x}_{i'}^{(j')}}z_{r,i,i'}^{(j,j')}
\end{align*}
with
\begin{align*}
    z_{r,i,i'}^{(j,j')} = \indy{\hat{\x}_i^{(j)};\w_{r,t}}\indy{\hat{\x}_{i'}^{(j')};\w_{r,t}} - \indy{\hat{\x}_i^{(j)};\w_{r,0}}\indy{\hat{\x}_{i'}^{(j')};\w_{r,0}}
\end{align*}
We observe that $|z_{r,i,i'}^{(j,j')}|$ only if $A_{ir}^{(j)}\vee A_{i'r}^{(j')}$. Therefore
\begin{align*}
    \E_{\mathbf{W}}\left[z_{r,i,i'}^{(j,j')}\right]\leq \mathbb{P}\paren{A_{ir}^{(j)}} + \mathbb{P}\paren{A_{i'r}^{(j')}}\leq \frac{4R}{\kappa\sqrt{2\pi}}
\end{align*}
For the case $j = j'$, we first notice that
\begin{align*}
    \E_{\mathbf{W}}\left[\paren{z_{r,i,i'}^{(j,j')} - \E_{\mathbf{W}}\left[z_{r,i,i'}^{(j,j')}\right]}^2\right] \leq \E_{\mathbf{W}}\left[z_{r,i,i'}^{(j,j')2}\right] \leq \frac{4R}{\kappa\sqrt{2\pi}}
\end{align*}
Thus, applying Berstein Inequality to the case $j = j'$ we have that
\begin{align*}
    \mathbb{P}\paren{\sum_{r=1}^mz_{r,i,i'}^{(j,j)}\geq m\paren{\E_{\mathbf{W}}\left[z_{r,i,i'}^{(j,j')}\right] + t}} \leq \exp\paren{-\frac{\kappa mt^2\sqrt{2\pi}}{8\paren{1 + \frac{t}{3}}R}}
\end{align*}
For the case $j\neq j'$, we notice that
\begin{align*}
    \E_{\W,\mathbf{a}}\left[a_{rj}a_{rj'}z_{r,i,i'}^{(j,j')}\right] = 0
\end{align*}
Moreover,
\begin{align*}
    \left|a_{rj}a_{rj'}z_{r,i,i'}^{(j,j')}\right|& \leq \frac{4R}{p^2m\kappa\sqrt{2\pi}}\\
    \E_{\W,\mathbf{a}}\left[\paren{a_{rj}a_{rj'}z_{r,i,i'}^{(j,j')}}^2\right] & =\frac{1}{p^4m^2}\E_{\W}\left[z_{r,i,i'}^{(j,j')2}\right] \leq\frac{4R}{p^4m^2\kappa\sqrt{2\pi}}
\end{align*}
Applying Berstein Inequality to the case $j\neq j'$, we have that
\begin{align*}
    \mathbb{P}\paren{\sum_{r=1}^ma_{rj}a_{rj'}z_{r,i,i'}^{(j,j')}\geq \frac{t}{p^2}} \leq \exp\paren{-\frac{m\kappa t^2\sqrt{2\pi}}{8\paren{1 + \frac{t}{3}}R}}
\end{align*}
Combining both cases, we have that with probability at least $1 - p^2\exp\paren{-\frac{m\kappa t^2\sqrt{2\pi}}{8\paren{1 + \frac{t}{3}}R}}$, it holds that
\begin{align*}
    \left|\h(t)_{i,i'} - \h(0)_{i,i'}\right|\leq p^{-1}\E_{\mathbf{W}}\left[z_{r,i,i'}^{(j,j')}\right] + t \leq \frac{2R}{p\kappa} + t^2
\end{align*}
Choose $t = \kappa^{-1}R$. Then as long as $m = \frac{\log \frac{np}{\delta}}{R}$, it holds that with probability at least $1 - O(\delta)$
\begin{align*}
    \left|\h(t)_{i,i'} - \h(0)_{i,i'}\right|\leq 3\kappa^{-1}R
\end{align*}
This implies that
\begin{align*}
    \norm{\h(t) - \h(0)}_2\leq\norm{\h(t)-\h(0)}_F \leq 3n\kappa^{-1}R
\end{align*}
Thus, $\norm{\h(t) - \h(0)}_2\leq \frac{\lambda_0}{4}$ as long as $R = O\paren{\frac{\kappa \lambda_0}{n}}$. This shows that $\lambda_{\min}\paren{\h(t)}\geq \frac{\lambda_0}{2}$ for all $t$ with probability at least $1 - O(\delta)$.
\end{proof}
\end{lemma}
\subsection{Surrogate Gradient Bound}
As we see in previous section, the one-step \textsc{LoFT} scheme can be written as 
\begin{align*}
    \w_{r,t+1} = \w_{r,t} - \eta\mathbf{g}_{r,t};\quad \mathbf{g}_{r,t} = \sum_{i=1}^n\sum_{j=1}^pa_{rj}\paren{\tilde{u}_{r,t}^{(i)} - N_{r,t}^\perp y_i}\hat{\x}_i^{(j)}\indy{\hat{\x}_i^{(j)};\w_{r,t}}
\end{align*}
with $\tilde{u}_{r,t}^{(i)}$ defined as
\begin{align*}
    \tilde{u}_{r,t}^{(i)} = \frac{N_{r,t}^\perp}{N_{r,t}}\sum_{s=1}^Sm_{r,t}^{(s)}\hat{u}_t^{(s, i)} = \sum_{r'=1}^m\sum_{j=1}^p\underbrace{\paren{\frac{N_{r,t}^\perp}{N_{r,t}}\sum_{s=1}^Sm_{r,t}^{(s)}m_{r',t}^{(s)}}}_{\nu_{r,r',t}}a_{rj}\sigma\paren{\inner{\hat{\x}_i^{(j)}}{\w_{r,t}}}
\end{align*}
The mixing of the surrogate function $\tilde{u}_{r,t}^{(i)}$ can be bounded by
\begin{align*}
    \E_{\mathbf{M}_t}\left[\tilde{u}_{r,t}^{(i)}\right] &= \sum_{s=1}^S\sum_{r'=1}^m\sum_{j=1}^p\E_{\mathbf{M}_t}\left[m_{r,t}^{(s)}m_{r,t'}^{(s)}\cdot\frac{N_{r,t}^\perp}{N_{r,t}}\right]a_{r'j}\sigma\paren{\inner{\hat{\x}^{(j)}}{\w_{r,t'}}}\\
    & = \xi\theta\sum_{r'=1}^m\sum_{j=1}^pa_{rj}\sigma\paren{\inner{\hat{\x}^{(j)}}{\w_{r,t}}} + (1-\xi)\theta\sum_{j=1}^pa_{rj}\sigma\paren{\inner{\hat{\x}^{(j)}_i}{\w_{r,t}}}\\
    & = \theta u_t^{(i)} + (1-\xi)\theta\underbrace{\sum_{j=1}^pa_{rj}\sigma\paren{\inner{\hat{\x}^{(j)}_i}{\w_{r,t}}}}_{\hat{\epsilon}_{r,t}^{(i)}}
\end{align*}
Therefore,
\begin{align*}
    \E_{\mathbf{M}_t}\left[\mathbf{g}_{r,t}\right] & = \sum_{i=1}^n\sum_{j=1}^na_{rj}\E_{\mathbf{M}_t}\left[\tilde{u}_t^{(i)} - N_{r,t}^\perp y_i\right]\hat{\x}_i^{(j)}\indy{\hat{\x}_i^{(j)};\w_{r,t}}\\
    & = \xi^{-1}\theta\nabla_{\w_r}\mathcal{L}\paren{\W_t} + (1-\xi)\theta\underbrace{\sum_{i=1}^n\sum_{j'=1}^pa_{rj'}\hat{\epsilon}_{r,t}^{(i)}\hat{\x}_i^{(j)}\indy{\hat{\x}_i^{(j)};\w_{r,t}}}_{\boldsymbol{\epsilon}_{r,t}}
\end{align*}
Now, we have
\begin{align*}
    \left|\hat{\epsilon}_{r,t}^{(i)}\right|\leq \frac{1}{\sqrt{m}}\norm{\w_{r,t}}_2;\quad \norm{\boldsymbol{\epsilon}_{r,t}}_2 \leq \frac{n}{\sqrt{m}}\left|\hat{\epsilon}_{r,t}^{(i)}\right|\leq \frac{n}{m}\norm{\w_{r,t}}_2
\end{align*}
Moreover, we would like to investigate the norm and norm squared of the gradient. In particular, we first notice that, under the case of $N_{t,r}^\perp = 1$, we have
\begin{align*}
    \mathbf{g}_{r,t} = \xi^{-1}\nabla_{\w_r}\mathcal{L}\paren{\W_t} + \sum_{i=1}^n\sum_{j=1}^pa_{rj}\paren{\tilde{u}_{r,t}^{(i)} - u_{t}^{(i)}}\hat{\x}_i^{(j)}\indy{\hat{\x}_i^{(j)};\w_{r,t}}
\end{align*}
Thus, we are interested in $\norm{\tilde{\U}_{r,t} - \U_t}_2$. Following from previous work \citep{Liao2021MaskedNTK} (lemma 19, 20, and 21), we have that
\begin{align*}
    \E_{\mathbf{M}_t}\left[\nu_{r,r',t}\mid N_{r,t}^\perp = 1\right]\begin{cases}
    \xi & \text{ if }r\neq r'\\
    1 & \text{ if }r = r'
    \end{cases}
    \quad\quad 
    \text{Var}\paren{\nu_{r,r',t}\mid N_{r,t}^\perp=1} = \begin{cases}
    \frac{\theta-\xi^2}{S} & \text{ if }r\neq r'\\
    0 & \text{ if }r = r'
    \end{cases}
\end{align*}
Therefore
\begin{align*}
    \E_{\mathbf{M}_t} & \left[\norm{\tilde{\U}_{r,t} - \U_t}_2^2\mid N_{r,t}^\perp=1\right]\\ & = \sum_{i=1}^n\E_{\mathbf{M}_t}\left[\paren{\sum_{r'=1}^m\sum_{j=1}^p\paren{\nu_{r,r',t}-\xi}a_{rj}\sigma\paren{\inner{\hat{\x}_i^{(j)}}{\w_{r',t}}}}^2\mid N_{r,t}^\perp=1\right]\\
    & = \sum_{i=1}^n\sum_{r'=1}^m\E_{\mathbf{M}_t}\left[\paren{\nu_{r,r',t}-\xi}^2\mid N_{r,t}^\perp=1\right]\paren{\sum_{j=1}^pa_{rj}\sigma\paren{\inner{\hat{\x}_i^{(j)}}{\w_{r',t}}}}^2\\
    & \leq \frac{1}{mp}\sum_{i=1}^n\sum_{r'=1}^m\E_{\mathbf{M}_t}\left[\paren{\nu_{r,r',t}-\xi}^2\mid N_{r,t}^\perp=1\right]\sum_{j=1}^p\sigma\paren{\inner{\hat{\x}_i^{(j)}}{\w_{r',t}}}^2\\
    & \leq \frac{1}{mp}\sum_{i=1}^n\sum_{r'\neq r}\text{Var}\paren{\nu_{r,r',t}\mid N_{r,t}^\perp=1}\sum_{j=1}^p\sigma\paren{\inner{\hat{\x}_i^{(j)}}{\w_{r',t}}}^2 + \\
    &\quad\quad\quad \frac{1}{mp}\sum_{i=1}^n\E_{\mathbf{M}_t}\left[\paren{\nu_{r,r,t}-\xi}^2\mid N_{r,t}^\perp=1\right]\sum_{j=1}^p\sigma\paren{\inner{\hat{\x}_i^{(j)}}{\w_{r',t}}}^2\\
    & \leq \frac{1}{mpS}(\theta-\xi)\sum_{r'\neq r}\sum_{i=1}^n\sum_{j=1}^p\inner{\hat{\x}_i^{(j)}}{\w_{r',t}}^2 + \frac{1}{mp}(\theta-\xi^2)\sum_{i=1}^n\sum_{j=1}^p\inner{\hat{\x}_i^{(j)}}{\w_{r,t}}^2
\end{align*}
With high probabilty it holds that
\begin{align*}
    \sum_{r=1}^m\sum_{i=1}^n\sum_{j=1}^p\inner{\hat{\x}_i^{(j)}}{\w_{r,0}}^2 \leq 2mnp\kappa^2 - mnpR^2
\end{align*}
Thus, with sufficiently large $m$, the second term is always smaller than the first term, and we have
\begin{align*}
    \E_{\mathbf{M}_t}\left[\norm{\tilde{\U}_{r,t} - \U_t}_2^2\mid N_{r,t}^\perp=1\right] \leq 8n\kappa^2(\theta-\xi^2)S^{-1}
\end{align*}
Now, we can compute that
\begin{align*}
    \E_{\mathbf{M}_t}\left[\norm{\mathbf{g}_{r,t}}_2\mid N_{r,t}^\perp=1\right] & =  \E_{\mathbf{M}_t}\left[\norm{\sum_{i=1}^n\sum_{j=1}^pa_{rj}\paren{\tilde{u}_{r,t}^{(i)} - u_{t}^{(i)}}\hat{\x}_i^{(j)}\indy{\hat{\x}_i^{(j)};\w_{r,t}}}\mid N_{r,t}=1\right] + \\
    &\quad\quad\quad \xi^{-1}\theta\norm{\nabla_{\w_r}\mathcal{L}\paren{\W_t}}_2\\
    & = \sqrt{\frac{n}{m}}\E_{\mathbf{M}_t}\left[\norm{\tilde{\U}_{r,t} - \U_t}_2\mid N_{r,t}=1\right] + \theta\sqrt{\frac{n}{m}}\norm{\U_t - \y}_2\\
    & \leq n\kappa\sqrt{\frac{\theta-\xi^2}{mS}} + \theta\sqrt{\frac{n}{m}}\norm{\U_t - \y}_2
\end{align*}
And we know that $\mathbf{g}_{r,t} = 0$ when $N_{r,t}^\perp=0$. Therefore,
\begin{align*}
    \E_{\mathbf{M}_t}\left[\norm{\mathbf{g}_{r,t}}_2\right]\leq \theta n\kappa\sqrt{\frac{\theta-\xi^2}{mS}} + \theta^2\sqrt{\frac{n}{m}}\norm{\U_t - \y}_2
\end{align*}
Similarly
\begin{align*}
    \E_{\mathbf{M}_t}\left[\norm{\mathbf{g}_{r,t}}_2^2\mid N_{r,t}^\perp=1\right] & = \frac{2n}{m}\E_{\mathbf{M}_t}\left[\norm{\tilde{\U}_{r,t} - \U_t}_2^2\mid N_{r,t}=1\right] + \frac{2\theta^2n}{m}\norm{\U_t - \y}_2^2\\
    & \leq \frac{16n^2\kappa^2\paren{\theta-\xi^2}}{mS} + \frac{2\theta^2n}{m}\norm{\U_t - \y}_2^2
\end{align*}
and thus
\begin{align*}
    \E_{\mathbf{M}_t}\left[\norm{\mathbf{g}_{r,t}}_2^2\right]\leq \frac{16\theta n^2\kappa^2\paren{\theta-\xi^2}}{mS} + \frac{2\theta^3n}{m}\norm{\U_t - \y}_2^2
\end{align*}
\subsection{Step-wise Convergence}Consider
\begin{align*}
    \E_{\mathbf{M}_t}\left[\norm{\U_{t+1} - \y}_2^2\right] & = \norm{\U_t - \y}_2^2 - 2\inner{\U_t- \U_{t+1}}{\U_t - \y} + \norm{\U_t - \U_{t+1}}_2^2\\
    & = \norm{\U_t - \y}_2^2 - 2\inner{\mathbf{I}_{1,t} + \mathbf{I}_{2,t}}{\U_t - \y} + \norm{\U_t - \U_{t+1}}_2^2\\
    & \leq \norm{\U_t - \y}_2^2 - 2\inner{\mathbf{I}_{1,t}}{\U_t - \y} + 2\norm{\mathbf{I}_{2,t}}_2\norm{\U_t - \y}_2 + \norm{\U_t - \U_{t+1}}_2^2
\end{align*}
Let $P_{ij} = \left\{r\in[m]:\neg A_{ir}^{(j)}\right\}$. Here $\mathbf{I}_{1,t}$ and $\mathbf{I}_{2,t}$ are characterized as in previous work.
\begin{align*}
I_{1,t}^{(i)} & = \sum_{j=1}^p\sum_{r\in P_{ij}}a_{rj}\paren{\sigma\paren{\inner{\xij}{\w_{r,t}}} - \sigma\paren{\inner{\xij}{\w_{r,t+1}}}}\\
    I_{2,t}^{(i)} & = \sum_{j=1}^p\sum_{r\in P_{ij}^\perp}a_{rj}\paren{\sigma\paren{\inner{\xij}{\w_{r,t}}} - \sigma\paren{\inner{\xij}{\w_{r,t+1}}}}
\end{align*}

We first bound the magnitude of $\mathbf{I}_{2,t}$
\begin{align*}
    \left|I_{2,t}^{(i)}\right| & = \frac{1}{p\sqrt{m}}\sum_{j=1}^p\sum_{r\in P_{ij}^\perp}\left|\paren{\sigma\paren{\inner{\hat{\x}_i^{(j)}}{\w_{r,t}}}-\sigma\paren{\inner{\hat{\x}_i^{(j)}}{\w_{r,t+1}}}}\right|\\
    & \leq \frac{1}{p\sqrt{m}}\sum_{j=1}^p\sum_{r\in P_{ij}}\norm{\w_{r,t} - \w_{r,t+1}}_2\\
    & \leq \frac{\eta}{p\sqrt{m}} \sum_{j=1}^p\sum_{r\in P_{ij}}\norm{\mathbf{g}_{r,t}}_2\\
    & \leq \frac{\eta}{\sqrt{m}}\cdot 3m\kappa^{-1}R\cdot \paren{n\kappa\sqrt{\frac{\theta-\xi^2}{mS}} + \theta\sqrt{\frac{n}{m}}\norm{\U_t - \y}_2}\\
    & = 3\kappa^{-1}\theta\eta\sqrt{n}R\norm{\U_t -\y}_2 + 3\eta nR\sqrt{\frac{\theta-\xi^2}{S}}
\end{align*}
Thus
\begin{align*}
    \E_{\mathbf{M}_t}\left[\norm{\mathbf{I}_{2,t}}_2\right] & \leq \E_{\mathbf{M}_t}\left[\sqrt{n}\max_{i\in[n]}\left|I_{2,t}^{(i)}\right|\right]\\
    & \leq \frac{\eta}{p} \sqrt{\frac{n}{m}}\max_{i\in[n]}\sum_{j=1}^p\sum_{r\in P_{ij}}\E_{\mathbf{M}_t}\left[\norm{\mathbf{g}_{r,t}}_2\right]\\
    & \leq \eta\sqrt{\frac{n}{m}}\cdot 3m\kappa^{-1}R\cdot \paren{n\kappa\theta\sqrt{\frac{\theta-\xi^2}{mS}} + \theta^2\sqrt{\frac{n}{m}}\norm{\U_t - \y}_2}\\
    & = 3\kappa^{-1}\theta^2\eta nR\norm{\U_t -\y}_2 + 3\theta\eta n^{\frac{3}{2}}R\sqrt{\frac{\theta-\xi^2}{S}}
\end{align*}
Therefore,
\begin{align*}
    \E_{\mathbf{M}_t}\left[\norm{\mathbf{I}_{2,t}}_2\norm{\U_t - \y}_2\right] \leq 6\theta\eta n\kappa^{-1}R\norm{\U_t - \y}_2^2 + 3S^{-1}\theta\paren{\theta-\xi^2}\eta n^2\kappa R
\end{align*}
Letting $R = O\paren{\frac{\kappa\lambda_0}{n}}$ gives that
\begin{align*}
    \E_{\mathbf{M}_t}\left[\norm{\mathbf{I}_{2,t}}_2\norm{\U_t - \y}_2\right] = O\paren{\theta\eta\lambda_0 }\norm{\U_t - \y}_2^2 + O\paren{\theta\eta\lambda_0(\theta-\xi^2)S^{-1}n\kappa^2}
\end{align*}

As in previous work, $I_{1,t}^{(i)}$ can be written as
\begin{align*}
    \E_{\mathbf{M}_t}\left[I_{1,t}^{(i)}\right] & = \xi\sum_{j=1}^p\sum_{r\in P_{ij}}a_{rj}\E_{\mathbf{M}_t}\left[\sigma\paren{\inner{\hat{\x}_i^{(j)}}{\w_{r,t}}}-\sigma\paren{\inner{\hat{\x}_i^{(j)}}{\w_{r,t+1}}}\right]\\
    & = \xi\sum_{j=1}^p\sum_{r\in P_{ij}}a_{rj}\inner{\hat{\x}_i^{(j)}}{\E_{\mathbf{M}_t}\left[\w_{r,t} - \w_{r,t+1}\right]}\indy{\hat{\x}_i^{(j)};\w_{r,t}}\\
    & = \xi\eta\sum_{j=1}^p\sum_{r\in P_{ij}}a_{rj}\inner{\hat{\x}_i^{(j)}}{\E_{\mathbf{M}_t}\left[\mathbf{g}_{r,t}\right]}\indy{\hat{\x}_i^{(j)};\w_{r,t}}\\
    & = \eta\theta\sum_{j=1}^p\sum_{r\in P_{ij}}a_{rj}\inner{\hat{\x}_i^{(j)}}{\nabla_{\w_r}\mathcal{L}\paren{\W_t}}\indy{\hat{\x}_i^{(j)};\w_{r,t}} +\\
    &\quad\quad\quad\xi(1-\xi)\theta\eta\sum_{j=1}^p\sum_{r\in P_{ij}}a_{rj}\inner{\hat{\x}_i^{(j)}}{\boldsymbol{\epsilon}_{r,t}}\indy{\hat{\x}_i^{(j)};\w_{r,t}}\\
    & = \eta\theta\xi\sum_{i'=1}^n\sum_{j,j'=1}^p\sum_{r\in P_{ij}}\paren{u_t^{(i')} - y_{i'}}a_{rj}a_{rj'}\inner{\hat{\x}_i^{(j)}}{\hat{\x}_{i'}^{(j')}}\indy{\hat{\x}_i^{(j)};\w_{r,t}}\cdot\\
    &\quad\quad\quad\indy{\hat{\x}_{i'}^{(j')};\w_{r,t}} + \xi(1-\xi)\theta\eta\sum_{j=1}^p\sum_{r\in P_{ij}}a_{rj}\inner{\hat{\x}_i^{(j)}}{\boldsymbol{\epsilon}_{r,t}}\indy{\hat{\x}_i^{(j)};\w_{r,t}}\\
    & = \eta\theta\sum_{i'=1}^n\paren{\h(t)_{ii'} - \h(t)^\perp_{ii'}}\paren{u_t^{(i')}-y_{i'}} +\\
    & \quad\quad\quad\underbrace{\xi(1-\xi)\theta\eta\sum_{j=1}^p\sum_{r\in P_{ij}}a_{rj}\inner{\hat{\x}_i^{(j)}}{\boldsymbol{\epsilon}_{r,t}}\indy{\hat{\x}_i^{(j)};\w_{r,t}}}_{\gamma_{i,t}}\\
\end{align*}
Note that
\begin{align*}
    \left|\gamma_{i,t}\right| & \leq \xi(1-\xi)\theta\eta m^{-\frac{1}{2}}\sum_{r=1}^m\norm{\boldsymbol{\epsilon}_{r,t}} \leq \xi(1-\xi)\theta\eta nm^{-1}\norm{\W_{t}}_F \leq O\paren{\xi(1-\xi)\theta\eta n\kappa\sqrt{\frac{d}{m}}}
\end{align*}
This implies that
\begin{align*}
    \E_{\mathbf{M}_t}\left[\inner{\mathbf{I}_{1,t}}{\U_t - \y}\right] & = \eta\theta\sum_{i,i'=1}^n\paren{u_t^{(i)}-y_{i}}\paren{\h(t)_{ii'} - \h(t)^\perp_{ii'}}\paren{u_t^{(i')}-y_{i'}} +\\
    &\quad\quad\quad\sum_{i=1}^n\gamma_{i,t}\paren{u_t^{(i)}-y_i}\\
    & = \eta\theta\inner{\U_t - \y}{\paren{\h(t) - \h(t)^\perp}\paren{\U_t - \y}} +\gamma_{r,t}\paren{u_t^{(i)}-y_i}\\
    & \geq \eta\theta\paren{\lambda_{\min}\paren{\h(t)} - \lambda_{\max}\paren{\h(t)^\perp}}\norm{\U_t - \y}_2^2 -\\
    &\quad\quad\quad \sum_{i=1}^n\left|\gamma_{i,t}\right|\cdot\left|u_t^{(i)} - y_i\right|\\
    & \geq \eta\theta\paren{\lambda_{\min}\paren{\h(t)} - \lambda_{\max}\paren{\h(t)^\perp}}\norm{\U_t - \y}_2^2 -\\
    &\quad\quad\quad \sqrt{n}\max_{i}\left|\gamma_{i,t}\right|\norm{\U_t - \y}_2\\
    & \geq \eta\theta\paren{\lambda_{\min}\paren{\h(t)} - \lambda_{\max}\paren{\h(t)^\perp} - O\paren{\lambda_0}}\norm{\U_t - \y}_2^2 -\\
    &\quad\quad\quad O\paren{\frac{\xi^2(1-\xi)^2\theta\eta n^3\kappa^2d}{m\lambda_0}}
\end{align*}
For $\h(t)^\perp$ we have that
\begin{align*}
    \lambda_{\max}\paren{\h(t)^\perp}^2 & \leq \norm{\h(t)^\perp}_F^2\\
    & \leq \sum_{i,i'=1}^n\paren{\sum_{j,j'=1}^p\sum_{r\in P_{ij}}a_{rj}a_{rj'}\inner{\xij}{\hat{\x}_{i,i'}^{(j,j')}}\indy{\xij;\w_{r,t}}\indy{\hat{\x}_{i,i'}^{(j,j')};\w_{r,t}}}^2\\
    & \leq \frac{n^2}{m^2}\paren{\max_{ij}\left|P_{ij}\right|}^2\\
    & \leq n^2\kappa^{-2} R^2
\end{align*}
Choosing $R = O\paren{\frac{\kappa\lambda_0}{n}}$ gives
\begin{align*}
    \lambda_{\max}\paren{\h(t)^\perp} \leq O\paren{\lambda_0}
\end{align*}
Plugging in $\lambda_{\min}\paren{\h(t)} \geq \frac{\lambda_0}{2}$, we have
\begin{align*}
    \E_{\mathbf{M}_t}\left[\inner{\mathbf{I}_{1,t}}{\U_t - \y}\right] & \geq \eta\theta\lambda_0\paren{\frac{1}{2} - O\paren{1}}\norm{\U_t - \y}_2^2 -O\paren{\frac{\xi^2(1-\xi)^2\theta\eta n^3\kappa^2d}{m\lambda_0}}
\end{align*}
Lastly, we analyze the last term in the quadratic expansion
\begin{align*}
    \E_{\mathbf{M}_t}\left[\norm{\U_t - \U_{t+1}}_2^2\right] & = \sum_{i=1}^n\E_{\mathbf{M}_t}\left[\paren{u_t^{(i)} - u_{t+1}^{(i)}}^2\right]\\
    & \leq p^{-1}\sum_{i=1}^n\sum_{j=1}^p\sum_{r=1}^m\E_{\mathbf{M}_t}\left[\paren{\sigma\paren{\inner{\xij}{\w_{r,t}}} - \sigma\paren{\inner{\xij}{\w_{r,t+1}}}}^2\right]\\
    & \leq p^{-1}\eta^2\sum_{i=1}^n\sum_{j=1}^p\sum_{r=1}^m\E_{\mathbf{M}_t}\left[\norm{\mathbf{g}_{r,t}}_2^2\right]\\
    & \leq O\paren{\theta^3\eta^2n^2}\norm{\U_t - \y}_2^2 + O\paren{\theta\eta^2n^2\kappa^2\paren{\theta-\xi^2}S^{-1}}
\end{align*}
Letting $\eta = O\paren{\frac{\lambda_0}{n^2}}$ gives
\begin{align*}
    \E_{\mathbf{M}_t}\left[\norm{\U_t - \U_{t+1}}_2^2\right] \leq O\paren{\theta\eta\lambda_0}\norm{\U_t - \y}_2^2 + O\paren{\theta\eta\lambda_0\kappa^2\paren{\theta-\xi^2}S^{-1}}
\end{align*}Putting all three terms together we have that
\begin{align*}
    \E_{\mathbf{M}_t}\left[\norm{\U_{t+1} - \y}_2^2\right] & \leq \paren{1 - \eta\theta\lambda_0\paren{1 - O\paren{1}}}\norm{\U_t - \y}_2^2 + O\paren{\frac{\xi^2(1-\xi)^2\theta\eta n^3\kappa^2d}{m\lambda_0}} + \\
    &\quad\quad\quad O\paren{ \theta\eta\lambda_0(\theta-\xi^2)S^{-1} n\kappa^2} +  O\paren{\theta\eta\lambda_0 \kappa^2\paren{\theta-\xi^2}S^{-1}}
\end{align*}
For sufficiently small constant in the upper bound of $R$, we have that
\begin{align*}
    \E_{\mathbf{M}_t}\left[\norm{\U_{t+1} - \y}_2^2\right] & \leq \paren{1 - \frac{\theta\eta\lambda_0}{2}}\norm{\U_t- \y}_2^2 + \eta\theta\lambda_0O\paren{\frac{\xi^2(1-\xi)^2 n^3d}{m\lambda_0^2} + \frac{n\kappa^2\paren{\theta-\xi^2}}{S}}
\end{align*}
Thus, we have that
\begin{align*}
    \E_{[\mathbf{M}_{t-1}]}\left[\norm{\U_{t} - \y}_2^2\right] & \leq \paren{1 - \frac{\theta\eta\lambda_0}{2}}^t\norm{\U_0 - \y}_2^2 + O\paren{\frac{\xi^2(1-\xi)^2 n^3d}{m\lambda_0^2} + \frac{n\kappa^2\paren{\theta-\xi^2}}{S}}
\end{align*}
\subsection{Bounding Weight Perturbation}
Next we show that $\norm{\w_{r,t} - \w_{r,0}}_2\leq R$ under sufficient over-parameterization. To start, we notice that
\begin{align*}
    \E_{[\mathbf{M}_{t-1}]}\left[\norm{\w_{r,t}-\w_{r,0}}_2\right] & \leq \sum_{t'=0}^{t-1}\E_{[\mathbf{M}_{t'}]}\left[\norm{\w_{r,t'+1} - \w_{r,t'}}_2\right]\\
    & \leq \eta\sum_{t'=0}^{t-1}\E_{[\mathbf{M}_{t'}]}\left[\norm{\mathbf{g}_{r,t'}}_2\right]\\
    & \leq \eta\sum_{t'=0}^{t-1}\paren{\theta^2\sqrt{\frac{n}{m}}\E_{[\mathbf{M}_{t'-1}]}\left[\norm{\U_{t'} - \y}_2\right] + \theta n\kappa\sqrt{\frac{\theta-\xi^2}{mS}}}\\
    & \leq \eta \theta^2\sqrt{\frac{n}{m}}\sum_{t'=0}^{t-1}\E_{[\mathbf{M}_{t'-1}]}\left[\norm{\U_{t'} - \y}_2\right] + \eta t \theta n\kappa\sqrt{\frac{\theta-\xi^2}{mS}}\\
    & \leq \eta\theta \sqrt{\frac{n}{m}}\norm{\U_0 - \y}_2\sum_{t'=0}^{t-1}\paren{1 - \frac{\eta\theta\lambda_0}{4}}^{t'} + \eta T \theta  n\kappa\sqrt{\frac{\theta-\xi^2}{mS}} + \\
    &\quad\quad\quad \eta\theta T\cdot O\paren{\frac{\xi(1-\xi)n^2\sqrt{d}}{m\lambda_0} + n\kappa\sqrt{\frac{\theta-\xi^2}{mS}}}\\
    & \leq O\paren{\lambda_0^{-1}\sqrt{\frac{n}{m}}}\norm{\U_0 - \y}_2 + \\
    &\quad\quad\quad \eta\theta T\cdot O\paren{\frac{\xi(1-\xi)n^2\sqrt{d}}{m\lambda_0} + n\kappa\sqrt{\frac{\theta-\xi^2}{mS}}}\\
\end{align*}
where the last inequality follows from the geometric sum and $\beta \leq O\paren{p^{-1}}$. Using the initialization scale, we have that
\begin{align*}
    \E_{\W,\mathbf{a},[\mathbf{M}_t]}\left[\norm{\w_{r,t}-\w_{r,0}}_2\right] \leq O\paren{\lambda_0^{-1} nm^{-\frac{1}{2}}} + \eta\theta  T\cdot O\paren{\frac{\xi(1-\xi)n^2\sqrt{d}}{m\lambda_0} + n\kappa\sqrt{\frac{\theta-\xi^2}{mS}}}
\end{align*}
With probability $1-\delta$, it holds for all $t\in [T]$ that
\begin{align*}
    \norm{\w_{r,t}-\w_{r,0}}_2 \leq O\paren{\frac{nK}{\lambda\delta\sqrt{m}}} + \eta\theta T\cdot O\paren{\frac{\xi(1-\xi)n^2K\sqrt{d}}{m\delta\lambda_0} + nK\kappa\sqrt{\frac{\theta-\xi^2}{mS\delta}}}
\end{align*}
To enforce $\norm{\w_{r,t}-\w_{r,0}}_2 \leq R := O\paren{\frac{\kappa\lambda_0}{n}}$, we then require
\begin{align*}
    m = \Omega\paren{\frac{n^3K^2}{\lambda_0^4\delta^2\kappa^2}\max\{n, d\}}
\end{align*}
\section{Proof of Theorem 1 in Main Text}
\label{main_proof}
To start, we consider \textsc{LoFT} with one local training step. By the definition of the masks, each filter is included in one and only one subnetwork. We consider the set of weights $\{\W_t\}$ training using \textsc{LoFT} and the set of weights $\{\hat{\W}_t\}$ trained using regular gradient descent
\begin{align*}
    \w_{r,t+1} = \w_{r,t} - \eta\frac{N_{r,t}^\perp}{N_{r,t}}\sum_{s=1}^S\nabla_{\w_r}\mathcal{L}_{\mathbf{m}_t^{(s)}}\paren{\W_{r,t}};\quad\hat{\w}_{r,t+1} = \hat{\w}_{r,t} - \eta\xi^{-1}\theta\nabla_{\w_r}\mathcal{L}\paren{\hat{\W}_t}
\end{align*}
From the last section, we know that with probability at least $1 - O(\delta)$, it holds that
\begin{align*}
    \norm{\w_{r,t} - \w_{r,0}}_2 \leq O\paren{\frac{n}{\lambda_0\sqrt{m}}}
\end{align*}
Also, notice that the iterates $\{\hat{\W}_t\}$ is the same as \textsc{LoFT} when $S = \xi = 1$. So we also have
\begin{align*}
    \norm{\hat{\w}_{r,t} - \hat{\w}_{r,0}}_2 \leq O\paren{\frac{n}{\lambda_0\sqrt{m}}}
\end{align*}
Therefore, naively we have that
\begin{align*}
    \norm{\w_{r,t} - \hat{\w}_{r,t}}_2 \leq O\paren{\frac{n}{\lambda_0\sqrt{m}}}
\end{align*}
Therefore, we can write $R:= O\paren{\frac{n}{\lambda_0\sqrt{m}}}  = O\paren{\frac{\kappa\lambda_0}{n}}$ under sufficient overparamterization.
for sufficient overparamterization.
The scaling here is for mathematical convenience in our analysis.
We start with expanding the squared difference of the two set of weights in iteration $t+1$
\begin{align*}
    \norm{\w_{r,t+1} - \hat{\w}_{r,t+1}}_2^2 & = \norm{\w_{r,t} - \hat{\w}_{r,t}}_2^2 + 2\inner{\w_{r,t} - \hat{\w}_{r,t}}{\w_{r,t+1} - \hat{\w}_{r,t+1} - \w_{r,t} + \hat{\w}_{r,t}} + \\
    & \quad\quad\quad\norm{\w_{r,t+1} - \hat{\w}_{r,t+1} - \w_{r,t} + \hat{\w}_{r,t}}_2^2\\
    & = \norm{\w_{r,t} - \hat{\w}_{r,t}}_2^2 - 2\eta\inner{\w_{r,t} - \hat{\w}_{r,t}}{\mathbf{g}_{r,t} - \theta\nabla_{\w_r}\mathcal{L}\paren{\hat{\W}_t}} +\\
    &\quad\quad\quad\eta^2\norm{\mathbf{g}_{r,t} - \theta\nabla_{\w_r}\mathcal{L}\paren{\hat{\W}_t}}_2^2
\end{align*}
Therefore
\begin{align*}
    \norm{\W_{t+1} - \hat{\W}_{t+1}}_F^2 & = \norm{\W_t - \hat{\W}_t}_F^2 - 2\eta\underbrace{\sum_{r=1}^m\inner{\w_{r,t} - \hat{\w}_{r,t}}{\mathbf{g}_{r,t} - \xi^{-1}\theta\nabla_{\w_r}\mathcal{L}\paren{\hat{\W}_t}}}_{Q_1} +\\
    &\quad\quad\quad\eta^2\underbrace{\sum_{r=1}^m\norm{\mathbf{g}_{r,t} - \xi^{-1}\theta\nabla_{\w_r}\mathcal{L}\paren{\hat{\W}_t}}_2^2}_{Q_2}
\end{align*}
To trace the dynamic of $\norm{\W_{t+1} - \hat{\W}_{t+1}}$, we need to analyze the second term (inner product) and the third term (second-order of the gradient difference) on the right-hand side of the equation. Denote them as $\eta Q_1$ and $\eta^2Q_2$, respectively.

\subsection{Analysis of the Second Term $Q_1$}
In previous section, we have seen that
\begin{align*}
    \E_{\mathbf{M}_t}\left[\mathbf{g}_{r,t}\right] = \xi^{-1}\theta\nabla_{\w_r}\mathcal{L}\paren{\W_t} + (1-\xi)\theta\boldsymbol{\epsilon}_{r,t}
\end{align*}
Therefore,
\begin{align*}
    \E_{\mathbf{M}_t}&\left[\inner{\w_{r,t} - \hat{\w}_{r,t}}{\mathbf{g}_{r,t} - \xi^{-1}\theta\nabla_{\w_r}\mathcal{L}\paren{\hat{\W}_t}}\right] \\
    & = \xi^{-1}\theta\inner{\w_{r,t} - \hat{\w}_{r,t}}{\nabla_{\w_r}\mathcal{L}\paren{\W_t} - \nabla_{\w_r}\mathcal{L}\paren{\hat{\W}}_t} + \\
    &\quad(1-\xi)\theta\inner{\w_{r,t} - \hat{\w}_{r,t}}{\boldsymbol{\epsilon}_{r,t}}
\end{align*}
In previous section, we have
\begin{align*}
    \left|\hat{\epsilon}_{r,t}^{(i)}\right|\leq \frac{1}{\sqrt{m}}\norm{\w_{r,t}}_2;\quad \norm{\boldsymbol{\epsilon}_{r,t}}_2 \leq \frac{n}{\sqrt{m}}\left|\hat{\epsilon}_{r,t}^{(i)}\right|\leq \frac{n}{m}\norm{\w_{r,t}}_2
\end{align*}
Therefore
\begin{align*}
    \left|\sum_{r=1}^m\inner{\w_{r,t} - \hat{\w}_{r,t}}{\boldsymbol{\epsilon}_{r,t}}\right| \leq \sum_{r=1}^m\norm{\w_{r,t} - \hat{\w}_{r,t}}_2\norm{\boldsymbol{\epsilon}_{r,t}}_2 \leq \frac{n^2\kappa}{\lambda_0}\sqrt{\frac{d}{m}}
\end{align*}

Denote the last term as $\Delta_1$. For convenience, we denote $u_t^{(i)} = f\paren{\hat{\x}_i,\W_t}$ and $\bar{u}_t^{(i)} = f\paren{\hat{\x}_i,\hat{\W}_t}$. Moreover, we have that
\begin{align*}
    \nabla_{\w_r}\mathcal{L}&\paren{\W_t} - \nabla_{\w_r}\mathcal{L}\paren{\hat{\W}_t} \\
    & = \xi\sum_{i=1}^n\sum_{j=1}^pa_{rj}\x_i^{(j)}\paren{\paren{u_t^{(i)} - y_i}\indy{\hat{\x}_i^{(j)};\w_{r,t}} - \paren{\bar{u}_t^{(i)} - y_i}\indy{\hat{\x}_i^{(j)};\hat{\w}_{r,t}}}
\end{align*}
Our goal is to study the term
\begin{align*}
    \alpha_1 = \sum_{r=1}^m\inner{\w_{r,t} - \hat{\w}_{r,t}}{\nabla_{\w_r}\mathcal{L}\paren{\W_t} - \nabla_{\w_r}\mathcal{L}\paren{\hat{\W}_t}}
\end{align*}
And we have
\begin{align*}
    \alpha_1 & = \xi\sum_{r=1}^m\sum_{i=1}^n\sum_{j=1}^pa_{rj}\paren{\inner{\hat{\x}_i^{(j)}}{\w_{r,t}} - \inner{\hat{\x}_i^{(j)}}{ \hat{\w}_{r,t}}}\cdot\\
    &\quad\quad\quad\paren{\paren{u_t^{(i)} - y_i}\indy{\hat{\x}_i^{(j)};\w_{r,t}} - \paren{\bar{u}_t^{(i)} - y_i}\indy{\hat{\x}_i^{(j)};\hat{\w}_{r,t}}}
\end{align*}
We should notice that 
\begin{align*}
    \inner{\hat{\x}_i^{(j)}}{\w_{r,t}}\indy{\hat{\x}_i^{(j)};\w_{r,t}} = \sigma\paren{\inner{\hat{\x}_i^{(j)}}{\w_{r,t}}};\; \inner{\hat{\x}_i^{(j)}}{ \hat{\w}_{r,t}}\indy{\hat{\x}_i^{(j)};\hat{\w}_{r,t}} = \sigma\paren{\inner{\hat{\x}_i^{(j)}}{\hat{\w}_{r,t}}}
\end{align*}
Therefore,
\begin{align*}
    \alpha_1 & = \sum_{r=1}^m\sum_{i=1}^n\sum_{j=1}^pa_{rj}\paren{\sigma\paren{\inner{\x_i^{(j)}}{\w_{r,t}}} - \sigma\paren{\inner{\x_i^{(j)}}{ \hat{\w}_{r,t}}}}\paren{\xi u_t^{(i)} - \xi\bar{u}_t^{(i)}} + \iota\\
    & = \sum_{i=1}^n\paren{u_t^{(i)} - \bar{u}_t^{(i)}}^2 + \iota
\end{align*}
with
\begin{align*}
    \iota & = \xi\sum_{r=1}^m\sum_{i=1}^n\sum_{j=1}^pa_{rj}\paren{\indy{\hat{\x}_i^{(j)};\w_{r,t}} - \indy{\hat{\x}_i^{(j)};\hat{\w}_{r,t}}}\cdot\\
    &\quad\quad\quad\paren{\inner{\hat{\x}_i^{(j)}}{ \hat{\w}_{r,t}}\paren{u_t^{(i)} - y_i} - \inner{\hat{\x}_i^{(j)}}{\w_{r,t}}\paren{\bar{u}_t^{(i)} - y_i}}
\end{align*}
Recall the definition of $P_{ij}$, we then have that for all $r\in P_{ij}$
\begin{align*}
    \indy{\hat{\x}_i^{(j)};\w_{r,t}} = \indy{\hat{\x}_i^{(j)};\w_{r,0}} = \indy{\hat{\x}_i^{(j)};\hat{\w}_{r,t}}
\end{align*}
Therefore, for $r\in P_{i,j}^\perp$, we have that
\begin{align*}
    \paren{\indy{\hat{\x}_i^{(j)};\w_{r,t}} - \indy{\hat{\x}_i^{(j)};\hat{\w}_{r,t}}}\inner{\hat{\x}_i^{(j)}}{\hat{\w}_{r,t}} & = -\left|\inner{\hat{\x}_i^{(j)}}{\hat{\w}_{r,t}}\right|\\
    \paren{\indy{\hat{\x}_i^{(j)};\w_{r,t}} - \indy{\hat{\x}_i^{(j)};\hat{\w}_{r,t}}}\inner{\hat{\x}_i^{(j)}}{\w_{r,t}} & = \left|\inner{\hat{\x}_i^{(j)}}{\w_{r,t}}\right|
\end{align*}
Therefore
\begin{align*}
    \left|\iota\right| & = \xi\left|\sum_{i=1}^n\sum_{j=1}^p\sum_{r\in P_{i,j}}a_{rj}\paren{\left|\inner{\hat{\x}_i^{(j)}}{\hat{\w}_{r,t}}\right|\paren{u_t^{(i)} - y_i} + \left|\inner{\hat{\x}_i^{(j)}}{\w_{r,t}}\right|\paren{\bar{u}_t^{(i)} - y_i}}\right|\\
    & \leq \xi\sum_{i=1}^n\sum_{j=1}^p\paren{\left|u_t^{(i)} - y_i\right|\left|\sum_{r\in P_{ij}}a_{rj}\left|\inner{\hat{\x}_i^{(j)}}{\hat{\w}_{r,t}}\right|\right| + \left|\bar{u}_t^{(i)} - y_i\right|\left|\sum_{r\in P_{ij}}a_{rj}\left|\inner{\hat{\x}_i^{(j)}}{\w_{r,t}}\right|\right|}\\
    & \leq \xi\sum_{i=1}^n\sum_{j=1}^p\paren{\left|u_t^{(i)} - y_i\right| + \left|\bar{u}_t^{(i)} - y_i\right|}\left|\sum_{r\in P_{ij}}a_{rj}\left|\inner{\hat{\x}_i^{(j)}}{\w_{r,0}}\right|\right| + \\
    &\quad\quad\quad\frac{\xi R\sqrt{n}}{p\sqrt{m}}\left|P_{i,j}\right|\paren{\norm{\U_t - \y}_2 + \norm{\bar{\U}_t - \y}_2}\\
    & \leq \xi \sqrt{n}\paren{p\left|\sum_{r\in P_{ij}}a_{rj}\left|\inner{\hat{\x}_i^{(j)}}{\w_{r,0}}\right|\right| + \frac{1}{\kappa\sqrt{m}}}\paren{\norm{\U_t - \y}_2 + \norm{\bar{\U_t} - \y}_2} +\\
    &\quad\quad\quad\frac{\sqrt{n}}{\kappa\sqrt{m}}\paren{\norm{\U_t - \y}_2 + \norm{\bar{\U}_t - \y}_2}
\end{align*}
Note that $\left|P_{ij}\right|\leq 3m\kappa^{-1}R$. Now we consider two cases of $R$:

\textbf{Case 1}: $R\leq \frac{n}{\lambda_0m^{\frac{3}{4}}}$. Then $\left|P_{ij}\right| \leq \frac{3nm^{\frac{1}{3}}}{\lambda_0\kappa}$. Then with high probability we have
\begin{align*}
    \left|\sum_{r\in P_{ij}}a_{rj}\left|\inner{\hat{\x}_i^{(j)}}{\w_{r,0}}\right|\right| \leq \frac{1}{p\sqrt{m}}\cdot \left|P_{ij}\right|\cdot \norm{\w_{r,0}}_2 \leq \frac{3n\sqrt{d}}{\lambda_0\kappa pm^{\frac{1}{4}}}
\end{align*}

\textbf{Case 2}: $\frac{n}{\lambda_0m^{\frac{3}{4}}} \leq R \leq \frac{n}{\lambda_0\sqrt{m}}$. Then $\left|P_{ij}\right|\leq \frac{3n\sqrt{m}}{\lambda_0\kappa}$. In this since $\norm{\hat{\x}_i^{(j)}}_2 = 1$ for all $i,j$, we know that $\inner{\hat{\x}_i^{(j)}}{\w_{r,0}}$ is Gaussian. Thus $p\sqrt{m}a_{rj}\left|\inner{\hat{\x}_i^{(j)}}{\w_{r,0}}\right|$ is Gaussian. Apply Hoeffding's inequality
\begin{align*}
    \mathbb{P}\paren{\left|\sum_{r\in P_{ij}}a_{rj}\left|\inner{\hat{\x}_i^{(j)}}{\w_{r,0}}\right|\right| \geq \frac{\left|P_{ij}\right|}{p\sqrt{m}}t}\leq \exp\paren{-\left|P_{ij}\right|t^2}
\end{align*}for all $\kappa$. Thus, it holds with probability at least $1- O(\delta)$ that
\begin{align*}
    \left|\sum_{r=1}^ma_{rj}\left|\inner{\hat{\x}_i^{(j)}}{\w_{r,0}}\right|\right| \leq \frac{\sqrt{\left|P_{ij}\right|\log\frac{pn}{\delta}}}{p\sqrt{m}} \leq \frac{\sqrt{n\log \frac{pn}{\delta}}}{\sqrt{\lambda_0\kappa}m^{\frac{1}{4}}}
\end{align*}
Combining both cases, we have that with probability at least $1 - O(\delta)$ it holds that
\begin{align*}
    \left|\sum_{r=1}^ma_{rj}\left|\inner{\bar{\x}_i^{(j)}}{\w_{r,0}}\right|\right| \leq \frac{3n\sqrt{d}}{\lambda_0\kappa pm^{\frac{1}{4}}}
\end{align*}
Therefore, we have
\begin{align*}
    |\iota|\leq \frac{3\xi\sqrt{n^3d}}{\lambda_0\kappa m^{\frac{1}{4}}}\paren{\norm{\U_t - \y}_2 + \norm{\hat{\U}_t - \y}_2}
\end{align*}
Thus, the second term is bounded by
\begin{align*}
    Q_1 \leq -\norm{\U_t - \hat{\U}_t}_2^2 + \frac{3\xi\sqrt{n^3d}}{\lambda_0\kappa m^{\frac{1}{4}}}\paren{\norm{\U_t - \y}_2 + \norm{\hat{\U}_t - \y}_2} + \frac{n^2\kappa}{\lambda_0}\sqrt{\frac{d}{m}}
\end{align*}
\subsection{Analysis of the Third Term}
Notice that
\begin{align*}
    \mathbf{g}_{r,t} - \xi^{-1}\theta\nabla_{\w_r}\mathcal{L}\paren{\hat{\W}_t} = \sum_{i=1}^n\sum_{j=1}^p\paren{\underbrace{\paren{\tilde{u}_{r,t}^{(i)} - \theta u_t^{(i)}}}_{\Delta_{1,t}^{(i)}} - \underbrace{\paren{N_{r,t}^\perp - \theta}y_i}_{\Delta_{2,t}^{(i)}}}a_{rj}\hat{\x}_i^{(j)}\indy{\hat{x}_i^{(j)};\hat{\w}_{r,t}}
\end{align*}
Therefore,
\begin{align*}
    \norm{\mathbf{g}_{r,t} - \xi^{-1}\theta\nabla_{\w_r}\mathcal{L}\paren{\hat{\W}_t}}_2^2 & = \sum_{i.i'=1}^n\sum_{j,j'=1}^pa_{rj}a_{rj'}\inner{\hat{\x}_{i}^{(j)}}{\hat{\x}_{i'}^{(j')}}\paren{\Delta_{1,t}^{(i)} - \Delta_{2,t}^{(i)}}\paren{\Delta_{1,t}^{(i')} - \Delta_{2,t}^{(i')}}\\
    & \leq p^{-1}\sum_{i,i'=1}^n\sum_{j=1}^p\inner{\hat{\x}_{i}^{(j)}}{\hat{\x}_{i'}^{(j)}}\paren{\Delta_{1,t}^{(i)} - \Delta_{2,t}^{(i)}}\paren{\Delta_{1,t}^{(i')} - \Delta_{2,t}^{(i')}}
\end{align*}
For $i\neq i'$, we notice that
\begin{align*}
    \E_{\mathbf{M}_t}\left[\paren{\Delta_{1,t}^{(i)} - \Delta_{2,t}^{(i)}}\paren{\Delta_{1,t}^{(i')} - \Delta_{2,t}^{(i')}}\right] & = \E_{\mathbf{M}_t}\left[\Delta_{1,t}^{(i)} - \Delta_{2,t}^{(i)}\right]\E_{\mathbf{M}_t}\left[\Delta_{1,t}^{(i')} - \Delta_{2,t}^{(i')}\right]\\
    & =(1-\xi)^2\theta^2\hat{\epsilon}_{r,t}^{(i)}\hat{\epsilon}_{r,t}^{(i')}
\end{align*}
Therefore,we can write
\begin{align*}
    \E_{\mathbf{M}_t}\left[\norm{\mathbf{g}_{r,t} - \xi^{-1}\theta\nabla_{\w_r}\mathcal{L}\paren{\hat{\W}_t}}_2^2\right] & \leq \frac{(1-\xi)^2\theta^2}{mp}\sum_{i=1}^n\sum_{i'\neq i}\sum_{j=1}^p\inner{\hat{\x}_{i}^{(j)}}{\hat{\x}_{i'}^{(j)}}\hat{\epsilon}_{r,t}^{(i)}\hat{\epsilon}_{r,t}^{(i')} + \\
    & \quad\quad\quad \frac{1}{mp}\sum_{i=1}^n\sum_{j=1}^p\E_{\mathbf{M}_t}\left[\paren{\Delta_{1,t}^{(i)} - \Delta_{2,t}^{(i)}}^2\right]\\
    & \leq \frac{(1-\xi)^2\theta^2n^2}{m^2}\norm{\w_{r,t}}_2^2 + \\
    &\quad\quad\quad\frac{1}{pm}\sum_{i=1}^n\sum_{j=1}^p\E_{\mathbf{M}_t}\left[\paren{\Delta_{1,t}^{(i)} - \Delta_{2,t}^{(i)}}^2\right]
\end{align*}
Studying the second term above requires analyzing
\begin{align*}
    \E_{\mathbf{M}_t}\left[\Delta_{1,t}^{(i)2}\right];\quad \E_{\mathbf{M}_t}\left[\Delta_{2,t}^{(i)2}\right];\quad \E_{\mathbf{M}_t}\left[\Delta_{1,t}^{(i)}\Delta_{2,t}^{(i)}\right]
\end{align*}
First, we have that
\begin{align*}
    \E_{\mathbf{M}_t}\left[\Delta_{2,t}^{(i)2}\right] = \E_{\mathbf{M}_t}\left[N_{r,t}^\perp - 2\theta N_{r,t}^\perp + \theta^2\right]y_i^2 = \theta(1-\theta)y_i^2 \leq \theta(1-\theta)C^2
\end{align*}
Notice that, by our definition of $N_{r,t}^\perp$ and $\tilde{u}_t^{(i)}$, we have $N_{r,t}^\perp\tilde{u}_t^{(i)} = \tilde{u}_t^{(i)}$. Therefore,
\begin{align*}
    \E_{\mathbf{M}_t}\left[\Delta_{1,t}^{(i)}\Delta_{2,t}^{(i)}\right] & = \E_{\mathbf{M}_t}\left[\tilde{u}_{r,t}^{(i)} - \theta N_{r,t}^\perp u_t^{(i)} - \theta\tilde{u}_t^{(i)} + \theta^2 u_t^{(i)}\right]y_i\\
    & = \theta(1-\theta)\paren{u_t^{(i)} + (1-\xi)\hat{\epsilon}_t^{(i)}}y_i
\end{align*}
Note that $\tilde{u}_{r,t}^{(i)}$ has the form
\begin{align*}
    \tilde{u}_{r,t}^{(i)} = \sum_{r=1}^m\sum_{j=1}^pa_{rj}\paren{\frac{N_{r,t}^\perp}{N_{r,t}}\sum_{s=1}^Sm_{r,t}^{(s)}m_{r',t}^{(s)}}\sigma\paren{\inner{\hat{\x}_i^{(j)}}{\w_{r,t}}} = \sum_{r=1}^m\sum_{j=1}^pa_{rj}\nu_{r,r',t}\sigma\paren{\inner{\hat{\x}_i^{(j)}}{\w_{r,t}}}
\end{align*}
by defining $\nu_{r,r',t} = \frac{N_{r,t}^\perp}{N_{r,t}}\sum_{s=1}^Sm_{r,t}^{(s)}m_{r',t}^{(s)}$. Lastly, let $\hat{\nu}_{r,r',t} = \nu_{r,r',t} - \xi\theta$, then we have
\begin{align*}
    \E_{\mathbf{M}_t} & \left[\Delta_{1,t}^{(i)2}\right]\\
    & = \E_{\mathbf{M}_t}\left[\paren{\sum_{r'=1}^m\sum_{j=1}^pa_{r'j}\hat{\nu}_{r,r',t}\sigma\paren{\inner{\hat{\x}_i^{(j)}}{\w_{r',t}}}}^2\right]\\
    & \leq p\E_{\mathbf{M}_t}\left[\sum_{r_1=1}^m\sum_{r_2\neq r_1}\sum_{j=1}^pa_{r_1j}a_{r_2j}\hat{\nu}_{r,r_1,t}\hat{\nu}_{r,r',t}\sigma\paren{\inner{\hat{\x}_i^{(j)}}{\w_{r_1,t}}}\sigma\paren{\inner{\hat{\x}_i^{(j)}}{\w_{r_2,t}}}\right] + \\
    &\quad\quad\quad \frac{1}{mp}\E_{\mathbf{M}_t}\left[\sum_{r'=1}^m\sum_{j=1}^p\hat{\nu}_{r,r',t}^2\sigma\paren{\inner{\hat{x}_i^{(j)}}{\w_{r',t}}}^2\right]\\
    & = \frac{1}{mp}\sum_{r'=1}^m\sum_{j=1}^p\E_{\mathbf{M}_t}\left[\hat{\nu}_{r,r',t}^2\right]\sigma\paren{\inner{\hat{\x}_i^{(j)}}{\w_{r,t}}}^2\\
    & \leq \frac{1}{mp}\sum_{r'=1}^m\sum_{j=1}^p\E_{\mathbf{M}_t}\left[\hat{\nu}_{r,r',t}^2\right]\inner{\hat{\x}_i^{(j)}}{\w_{r,t}}^2
\end{align*}
We need several property of $\nu_{r,r',t}$
\begin{align*}
    \E_{\mathbf{M}_t}\left[\nu_{r,r',t}\right] = \begin{cases}
    \xi\theta & \text{ if } r\neq r'\\
    \theta & \text{ if } r = r'
    \end{cases}
    \quad\quad
    \E_{\mathbf{M}_t}\left[\nu_{r,r',t}^2\right] = \begin{cases}
    \xi^2\theta^2 + \frac{\theta^2(1-\xi)}{S} & \text{ if } r\neq r'\\
    \theta & \text{ if } r = r'
    \end{cases}
\end{align*}
Thus,
\begin{align*}
    \E_{\mathbf{M}_t}\left[\hat{\nu}_{r,r',t}^2\right] = \begin{cases}
    \frac{\theta^2(1-\xi)}{S} & \text{ if }r \neq r'\\
    \theta - 2\xi\theta^2 + \xi^2\theta^2 & \text{ if } r = r'
    \end{cases}
\end{align*}
and therefore
\begin{align*}
    \E_{\mathbf{M}_t}\left[\Delta_{1,t}^{(i)2}\right]\leq \frac{\theta^2(1-\xi)}{mS}\sum_{r'=1}^m\sum_{j=1}^p\inner{\hat{\x}_i^{(j)}}{\w_{r,t}}^2 + \frac{\theta - 2\xi\theta^2 + \xi^2\theta^2}{m}\norm{\w_{r,t}}_2^2 \leq \frac{2\theta^2(1-\xi)\kappa^2}{S}
\end{align*} for sufficiently large $m$.
Thus,
\begin{align*}
    \E_{\mathbf{M}_t}\left[\paren{\Delta_{1,t}^{(i)} - \Delta_{2,t}^{(i)}}^2\right] \leq \frac{2\theta^2(1-\xi)\kappa^2}{S} +  \theta(1-\theta)\paren{u_t^{(i)} - y_i}^2 + \frac{\theta(1-\xi)^2C}{\sqrt{m}}\norm{\w_{r,t}}_2
\end{align*}
Putting things together, we have that
\begin{align*}
    \E_{\mathbf{M}_t}\left[\norm{\mathbf{g}_{r,t} - \xi^{-1}\theta\nabla_{\w_r}\mathcal{L}\paren{\W_t}}_2^2\right] & \leq \frac{(1-\xi)^2\theta^2n^2}{m^2}\norm{\w_{r,t}}_2^2 + \frac{2\theta^2(1-\xi)\kappa^2 n}{S} +\\
    &\quad\quad\quad\frac{\theta(1-\theta)}{m}\norm{\U_t - \y}_2^2 + \frac{\theta(1-\xi)^2Cn}{m^{\frac{3}{2}}}\norm{\w_{r,t}}_2
\end{align*}
Therefore,
\begin{align*}
    \sum_{r=1}^m\E_{\mathbf{M}_t}\left[\norm{\mathbf{g}_{r,t} - \xi^{-1}\theta\nabla_{\w_r}\mathcal{L}\paren{\W_t}}_2^2\right] & \leq \frac{(1-\xi)^2\theta^2n^2\kappa^2d}{m} + \frac{2\theta^2(1-\xi)\kappa^2 n}{S} +\\
    & \quad\quad\quad \theta(1-\theta)\norm{\U_t - \y}_2^2 +  \frac{\theta(1-\xi)^2Cn\kappa\sqrt{d}}{\sqrt{m}}\\
    & \leq \frac{2\theta^2(1-\xi)\kappa^2 n}{S} + \theta(1-\theta)\norm{\U_t - \y}_2^2
\end{align*}
Moreover
\begin{align*}
    \nabla_{\w_r}\mathcal{L}\paren{\W_t} & - \nabla_{\w_r}\mathcal{L}\paren{\hat{\W_t}} \\ & = \frac{\xi}{p\sqrt{m}}\sum_{i=1}^n\sum_{j=1}^pa_{rj}\hat{\x}_i^{(j)}\paren{\paren{u_t^{(i)} - y_i}\indy{\hat{\x}_i^{(j)};\w_{r,t}} - \paren{\hat{u}_t^{(i)} - y_i}\indy{\hat{\x}_i^{(j)};\hat{\w}_{r,t}}}
\end{align*}
Thus
\begin{align*}
    \sum_{r=1}^m & \norm{\nabla_{\w_r}\mathcal{L}\paren{\W_t} - \nabla_{\w_r}\mathcal{L}\paren{\hat{\W_t}}}_2^2 
    \\ & \leq \frac{\xi^2n}{pm}\sum_{i=1}^n\sum_{j=1}^p\sum_{r=1}^m\left|\paren{u_t^{(i)} - y_i}\indy{\hat{\x}_i^{(j)};\w_{r,t}} - \paren{\bar{u}_t^{(i)} - y_i}\indy{\hat{\x}_i^{(j)};\hat{\w}_{r,t}}\right|^2\\
    &\leq \frac{\xi^2n}{pm}\sum_{i=1}^n\sum_{j=1}^p\sum_{r\in P_{ij}}\left|\paren{u_t^{(i)} - y_i}\indy{\hat{\x}_i^{(j)};\w_{r,t}} - \paren{\bar{u}_t^{(i)} - y_i}\indy{\hat{\x}_i^{(j)};\hat{\w}_{r,t}}\right|^2+\\
    &\quad\quad\quad
    \frac{\xi^2n}{pm}\sum_{i=1}^n\sum_{j=1}^p\sum_{r=1}^m\left|u_t^{(i)} - \bar{u}_t^{(i)}\right|^2\\
    & \leq 3\xi^2n\kappa^{-1}R\paren{\norm{\U_t -\y}_2^2 + \norm{\bar{\U}_t - \y}_2^2} + \xi^2n\norm{\U_t - \bar{\U}_t}_2^2\\
    & \leq \frac{3\xi^2n^2}{\kappa\lambda_0\sqrt{m}}\norm{\U_t -\y}_2^2 + 2\xi^2n\norm{\U_t - \bar{\U}_t}_2^2
\end{align*}
by plugging in $R \leq O\paren{\frac{n}{\lambda_0\sqrt{m}}}$.
Thus, the third term here is bounded by
\begin{align*}
    Q_2 \leq O\paren{\frac{2\theta^2(1-\xi)\kappa^2 n}{S} + \frac{3\xi^2n^2}{\kappa\lambda_0\sqrt{m}}\norm{\U_t -\y}_2^2 + 2\xi^2n\norm{\U_t - \bar{\U}_t}_2^2}
\end{align*}
Combining all three conditions, we have that
\begin{align*}
    \E_{\mathbf{M}_t}\left[\norm{\W_{t+1} -\hat{\W}_{t+1}}_2^2\right] & \leq \norm{\W_{t} -\hat{\W}_{t}}_2^2 - \eta\norm{\U_t - \bar{\U}_t}_2^2 +\\
    &\quad\quad\quad\frac{3\eta\xi\sqrt{n^3d}}{\lambda_0\kappa m^{\frac{1}{4}}}\paren{\norm{\U_t - \y}_2 + \norm{\bar{\U}_t - \y}_2} + \frac{\eta n^2\kappa}{\lambda_0}\sqrt{\frac{d}{m}} +\\
    &\quad\quad\quad\frac{2\eta^2\theta^2(1-\xi)\kappa^2 n}{S} + \frac{3\xi^2\eta^2n^2}{\kappa\lambda_0\sqrt{m}}\norm{\U_t -\y}_2^2 + 2\xi^2\eta^2n\norm{\U_t - \bar{\U}_t}_2^2\\
    & \leq \norm{\W_{t} -\hat{\W}_{t}}_2^2 - \frac{\eta}{2}\norm{\U_t - \bar{\U}_t}_2^2 +\\
    & \quad\quad\quad\frac{3\eta\xi\sqrt{n^3d}}{\lambda_0\kappa m^{\frac{1}{4}}}\paren{\norm{\U_t - \y}_2 + \norm{\bar{\U}_t - \y}_2} + \frac{\eta n^2\kappa}{\lambda_0}\sqrt{\frac{d}{m}} + \\
    &\quad\quad\quad\frac{2\eta^2\theta^2(1-\xi)\lambda_0\kappa^2n}{S} + \frac{3\xi^2\eta}{\kappa\sqrt{m}}\norm{\U_t -\y}_2^2\\
\end{align*}
by taking $\eta = O\paren{\frac{\lambda_0}{n^2}}$.
Let's first make some simplification. Notice that by choosing $\kappa = O\paren{\frac{1}{\sqrt{n}}}$, we have
\begin{align*}
    \E_{\mathbf{M}_t}\left[\norm{\U_t - \y}_2^2\right] \leq \paren{1 -\frac{\eta\theta\lambda_0}{2}}^t\norm{\U_0 - \y}_2^2 + O(1)\\
    \E_{\mathbf{M}_t}\left[\norm{\bar{\U}_t - \y}_2^2\right] \leq \paren{1 -\frac{\eta\theta\lambda_0}{2}}^t\norm{\U_0 - \y}_2^2
\end{align*}
Thus, taking the total expectation
\begin{align*}
    \E_{[\mathbf{M}_T]}\left[\norm{\W_T - \hat{\W}_T}_F^2\right] & \leq \norm{\W_0 - \W_0}_F^2 - \eta\sum_{t=0}^{T-1}\E_{[\mathbf{M}_T]}\left[\norm{\U_t - \bar{\U}_t}_2^2\right] +\\
    &\quad\quad\quad O\paren{\frac{\sqrt{n^3d}}{\lambda_0^2\kappa m^{\frac{1}{4}}}}\norm{\U_0 - \y}_2 +  O\paren{\frac{\xi}{\kappa\lambda_0\sqrt{m}}}\norm{\U_0 -\y}_2^2 + \\
    &\quad\quad\quad O\paren{\frac{\sqrt{n^3d}}{\lambda_0^2\kappa m^{\frac{1}{4}}} + \frac{2\eta^2T\theta^2(1-\xi)\lambda_0}{S}}\\
    & \leq -\eta\sum_{t=0}^{T-1}\E_{[\mathbf{M}_T]}\left[\norm{\U_t - \bar{\U}_t}_2^2\right] +\\
    &\quad\quad\quad O\paren{\frac{n^2\sqrt{d}}{\lambda_0^2\kappa m^{\frac{1}{4}}\sqrt{\delta}} + \frac{2\eta^2T\theta^2(1-\xi)\lambda_0}{S}} 
\end{align*}
This brings to the conclusion
\begin{align*}
    \E_{[\mathbf{M}_T]}\left[\norm{\W_T - \hat{\W}_T}_F^2\right] + \eta\sum_{t=0}^{T-1}\E_{[\mathbf{M}_T]}\left[\norm{\U_t - \bar{\U}_t}_2^2\right] \leq O\paren{\frac{n^2\sqrt{d}}{\lambda_0^2\kappa m^{\frac{1}{4}}\sqrt{\delta}} + \frac{2\eta^2T\theta^2(1-\xi)\lambda_0}{S}}
\end{align*}

\end{document}